\documentclass[a4paper,fleqn]{cas-sc}

\usepackage[authoryear,longnamesfirst]{natbib}
\usepackage{graphicx}
\usepackage{amsmath}
\usepackage{booktabs} % For \toprule, \midrule, \bottomrule
\usepackage{tabularx} % For adjustable width tables
\usepackage{cite}
\usepackage{float}
\usepackage{placeins}
\usepackage{amsmath,amssymb,amsfonts}
\usepackage{algorithm}
\usepackage{algorithmic}
\usepackage{textcomp}
\usepackage{natbib}
\usepackage{hyperref}% for hyperlinks
\usepackage{lipsum}

\def\tsc#1{\csdef{#1}{\textsc{\lowercase{#1}}\xspace}}
\tsc{WGM}
\tsc{QE}
\tsc{EP}
\tsc{PMS}
\tsc{BEC}
\tsc{DE}
\begin{document}

\title [mode = title]{A Two-Stage Ensemble Feature Selection and Particle Swarm  Optimization Approach for Micro-Array Data Classification in Distributed Computing Environments}                      
%\tnotemark[1,2]

\tnotetext[1]{This document is the results of the research
   project funded by (Add the funding institute/agency here).}

\tnotetext[2]{The second title footnote which is a longer text matter
   to fill through the whole text width and overflow into
   another line in the footnotes area of the first page.}

% Authors and ORCID links
% Authors and ORCID links
\author[1]{Aayush Adhikari\corref{equalcontrib}}  % Equal contribution note
\ead{ayushadhikari3466@gmail.com}

\author[1]{Sandesh Bhatta\corref{equalcontrib}}  % Equal contribution note
\ead{bhatta.sandesh2003@gmail.com}

\author[2]{Harendra S. Jangwan}  % Equal contribution note
\ead{harendrajangwan@gmail.com}

\author[3]{Amit Mishra}  % Equal contribution note
\ead{amit.mishra@MITWPU.edu.in}

\author[5]{Khair Ul Nisa }  % Equal contribution note
\ead{drkhairulnisa@gmail.com}

\author[4]{Abu Taha Zamani}  % Equal contribution note
\ead{abutaha.zamani@nbu.edu.sa}

\author[1]{Aaron Sapkota\corref{equalcontrib}}  % Equal contribution note
\ead{aaronsapkotanasa@gmail.com}

\author[1]{Debendra Muduli\corref{correspondingauthor}}  % Corresponding author note
\ead{muduli.debendra@gmail.com}

\author[5]{Nikhat Parveen}  % Equal contribution note
\ead{nikhat0891@gmail.com}

\cortext[correspondingauthor]{Corresponding author: Debendra Muduli}
\cortext[equalcontrib]{These authors contributed equally to this work.}

% Author affiliation
\affiliation[1]{organization={Department of Computer Science \& Engineering, C.V. Raman Global University, 752054, India}}
\affiliation[2]{Department of Computer Science \& Engineering Quantum University, Uttarakhand, India.}
\affiliation[3]{Department of Computer Science, Dr. Vishwanath Karad, MITWPU, Pune, Maharashtra, India.}
\affiliation[4]{Department of Computer Science at Northern Border University, Arar, 73213, Saudi Arabia.}
\affiliation[5]{Department of Artificial Intelligence, College of Computing and Information Technology, University of Bisha, 67711, Saudi Arabia}

\nonumnote{This note has no numbers. In this work................................
  }

\begin{abstract}
High dimensionality in datasets produced by microarray technology presents a challenge for Machine Learning (ML) algorithms, particularly in terms of dimensionality reduction and handling imbalanced sample sizes. To mitigate the explained problems, we have proposed hybrid ensemble feature selection techniques with majority voting classifier for micro array classification. Here we have considered both filter and wrapper-based feature selection techniques including Mutual Information (MI), Chi-Square, Variance Threshold (VT), Least Absolute Shrinkage and Selection Operator (LASSO), Analysis of Variance (ANOVA), and Recursive Feature Elimination (RFE), followed by Particle Swarm Optimization (PSO) for selecting the optimal features. This Artificial Intelligence (AI) approach leverages a Majority Voting Classifier that combines multiple machine learning models, such as Logistic Regression (LR), Random Forest (RF), and Extreme Gradient Boosting (XGBoost), to enhance overall performance and accuracy. By leveraging the strengths of each model, the ensemble approach aims to provide more reliable and effective diagnostic predictions. The efficacy of the proposed model has been tested in both local and cloud environments. In the cloud environment, three virtual machines virtual Central Processing Unit (vCPU) with size 8,16 and 64 bits, have been used to demonstrate the model performance. From the experiment it has been observed that, virtual Central Processing Unit (vCPU) -64 bits provides better classification accuracies of 95.89\%, 97.50\%, 99.13\%, 99.58\%, 99.11\%, and 94.60\% with six microarray datasets, Mixed Lineage Leukemia (MLL), Leukemia, Small Round Blue Cell Tumors (SRBCT), Lymphoma, Ovarian, and Lung, respectively, validating the effectiveness of the proposed model in both local and cloud environments. 
\end{abstract}

% \includegraphics{figs/grabs.pdf}

% Keywords

	\maketitle
	%============================= INTRODUCTION===============================
 
\section{Introduction}
Microarray technology has revolutionized bioinformatics by enabling the simultaneous measurement of thousands of gene expressions, significantly advancing biomarker discovery, disease diagnosis, and treatment planning. These datasets are crucial in biomedical research, particularly in cancer classification, where identifying gene expression patterns helps in early detection and personalized therapy. For instance, datasets like MLL and Leukemia 3C aid in leukemia subtype classification, while SRBCT and Lymphoma datasets assist in distinguishing pediatric tumors and lymphatic cancers. Similarly, Ovarian and Lung cancer datasets provide insights into tumor progression and potential therapeutic targets. Leveraging these datasets allows for more accurate disease classification, guiding clinicians in making targeted and effective treatment decisions \citep{huttenhower2006scalable, muduli2021enhancement}.

A microarray dataset, conventionally termed a gene expression dataset, generally encompasses extensive volumes of information, with each gene serving as a unique feature or dimension. These datasets are highly complex, as they contain a massive amount of data—each gene represents a separate piece of information. The elevated dimensionality of these datasets presents both an advantageous opportunity and a formidable challenge \citep{bolon2014review, muduli2023enhancing}. Although microarray datasets possess substantial potential for revealing novel biological insights, they also introduce considerable obstacles for machine learning algorithms owing to their intrinsic complexity.

One of the most challenging tasks associated with the analysis of microarray data is the phenomenon of the curse of dimensionality, where the number of features significantly exceeds the number of available samples. Typically, microarray experiments are constrained to a limited number of samples due to the financial and logistical complexities associated with data acquisition \citep{harvey2015cloud}. This results in a high-dimensional, low-sample-size problem wherein thousands of genes are assessed against merely a few dozen or a few hundred samples. The proliferation of features not only escalates computational expenses but also engenders noise and redundancy within the dataset, thereby complicating the identification of salient patterns \citep{muduli2020automated}.

Many genes present within these datasets contribute little to no biological insight and act as noise, further reducing the operational efficacy of machine learning models \citep{gamal2022ensemble}. Consequently, models may become overly complex, performing well on training data but failing to generalize to unseen data. Moreover, microarray datasets frequently exhibit class imbalance, where the distribution of samples across different classes is uneven—leading to biased learning and inaccurate predictions \citep{razzaque2024pca, muduli2022automated}.

To overcome these challenges, we present a two-stage ensemble feature selection approach combined with Particle Swarm Optimization (PSO) for dimensionality reduction and improved classification performance. This approach also emphasizes the importance of feature selection—a key preprocessing step for high-dimensional microarray data that preserves only the most relevant genes. We employ multiple filter-based methods such as Mutual Information, Chi-Square, ANOVA, LASSO, and Variance Threshold, along with Recursive Feature Elimination (RFE). Each method evaluates feature relevance differently: Chi-Square and ANOVA assess statistical dependencies, Mutual Information quantifies entropy-based relevance, while LASSO eliminates features by applying coefficient regularization.

To ensure this method is practical for large-scale biomedical applications, we implement the model in distributed computing environments, which provide crucial advantages for high-dimensional data processing. Distributed systems offer parallelization, scalability, and reduced runtime, making them ideal for real-time or cloud-integrated medical applications. Performing dimensionality reduction in distributed systems ensures timely execution and seamless deployment, even when handling extremely large datasets. Furthermore, integrating feature selection techniques within such systems enhances their efficiency, especially for resource-constrained healthcare setups.

Our working hypothesis is that combining filter-based methods with wrapper-based optimization in a distributed setup leads to:
\begin{itemize}

\item More robust and stable feature subsets,

\item Improved classification accuracy,

\item Faster runtime due to parallel processing, and

\item Increased real-world applicability through cloud integration.
\end{itemize}

We further enhance classification performance using a weighted voting classifier comprising XGBoost, Random Forest, and Logistic Regression. This comprehensive pipeline can be extended to other medical disciplines, aiming to support accurate diagnostics and prognostics via machine learning.

The main objectives of this work are as follows:
\begin{itemize}
\item To identify the best subset of features from high-dimensional microarray datasets to improve proposed model accuracy.
\item To propose an ensemble feature selection approach by using six filter methods together with one wrapper method, applying the rank of the chosen features using the PSO technique.
\item To develop a lightweight model optimized for low-bandwidth environments, ensuring efficient processing and accessibility without requiring extensive computational resources.
\item To implement a cloud-integrated architecture that facilitates scalable, real-time analysis and seamless deployment across diverse healthcare platforms, enhancing the model's applicability in practical settings.
\end{itemize}

The paper is organized as follows: Section \ref{sec:p1} provides an overview of the relevant literature and theoretical foundations for the study. In Section \ref{sec:p2}, we explain how we processed the data, selected features, and used optimization and classification methods, including in the cloud. Section \ref{sec:p3} describes our microarray classification model and the experiments we ran. Section \ref{sec:p4} shares the results and compares performance in the cloud and standalone setups. Section \ref{sec:p6} explains about the future scopes. Finally, Section \ref{sec:p5} concludes the paper.

\section{Literature Review}
\label{sec:p1}
Microarray technology enables large-scale gene expression analysis but presents challenges due to high dimensionality, class imbalance, and redundancy. To address these, we propose a two-stage ensemble feature selection framework integrating six filter methods, Recursive Feature Elimination (RFE), and Particle Swarm Optimization (PSO) for optimal gene selection. A weighted voting classifier combining XGBoost, Random Forest, and Logistic Regression enhances classification accuracy by leveraging their complementary strengths. Unlike previous methods, our approach is evaluated across standalone and cloud-based environments, demonstrating scalability and efficiency. Tested on six microarray datasets, our model achieves state-of-the-art accuracy, making it a robust solution for biomedical data classification with potential applications in disease diagnosis and precision medicine.\citep{chaudhuri2021hybrid}.

Canedo et al. \citep{bolon2014review} challenges of microarray data classification, which includes high dimensionality and class imbalance. Feature selection is critical for reducing dimensionality and it evaluates common feature selection 
appraoches using well known classifier. The method selection should be based on dataset characteristics and research objectives. Tripathy et al. \citep{tripathy2022combination} proposes a combination of Feature Reduction(CFR) approach using four different 
filter technique to reduce high dimensional microarray data. Different combination are evaluated with classifiers 
like Decision Tree, KNN (K-Nearest Neighbors), and Logistic Regression. The TOPSIS is used to rank and identify the best-performing filter 
combination.  Chaudhuri et al. \citep{chaudhuri2021hybrid} inroduced a hybrid feature selction for high dimensional 
microarray data. This model combines the filtering of key features by the TOPSIS method with the optimization of 
the feature subset via the Binary Jaya algorithm. This model was tested on 10 datasets, and it achieved better 
classification accuracy and ran 10 times faster than the standard model. This means that this is an effective solution for processing complex data. 

Gamal et al. \citep{gamal2022ensemble} introduced Single-Valued Neutrosophic Sets(SVNS) approach combining AHP (Analytic Hierarchy Process), TOPSIS, and VIKOR (VIseKriterijumska Optimizacija I Kompromisno Resenje) which handles uncertainty and incomplete information in acute leukemia evaluation.The method improves decision making by accurately determining criteria weights and ranking alternatives. Marwai et al. \citep{marwai2024hybrid} proposed feature selection method, combining TOPSIS for filtering and Particle Swarm Optimization(PSO) for feature selection, which effectively reduces the dimensionality of microarray data. When tested on ADNI (Alzheimer’s Disease Neuroimaging Initiative) datasets, the appraoch outperforms benchmark methods in terms of classification accuracy. Huttenhower et al. \citep{huttenhower2006scalable} used Microarray Experiment Functional Integration Technology (MEFIT) framework, applied to 40 Saccharomyces cerevisiae microarray datasets which improves the accuracy by 5\% by predicting functional relationships and incresead accuracy for 54 out of 110 biological functions. Alanni et al. \citep{alanni2019novel} finds that the gene selection is important to overcome the curse of dimensionality in microarray datasets which usually have large number of genes but samll sample sizes. Gene selection helps reduce noise and irrelevant data which enhances the efficiency and accuracy of cancer classification.

Harvey et al. \citep{harvey2015cloud} introduced a new cloud-based method for analyzing large-scale cancer genomic data using a wavelet technique. This technique cleans and refines data, enhances the accuracy. This shows that the cloud computing and advanced wavelet processing make the analysis faster and precise. Sharma et al. \citep{sharma2023diabetes} developed a cloud-based 
automated eHealth system for the early detection of diabetes. ELM was used for classification along with Principal 
Component Analysis (PCA) because the feature selection mechanism which it is working under cloud environment 
with numerous virtual machines having varying capacity. It has been applied for PIMA dataset, and in that process, good accuracy is achieved. Razzaque et al. \citep{razzaque2024pca} introduced MC-FE method for classifying genomic dataset. It utilized PCA for feature extraction and MPSO for selecting features. The approached was tested with SVM, KNN and, Naive Bayes which demonstrates optimization result. Cilia et al. \citep{cilia2019experimental} finds that the feature selection improves classification performance in DNA microaaray dataset but no any single method outperforms across all dataset. The effectiveness of feature selection depends on dataset characteristics. Proper feature selection is important for enhancing the classification tasks. Deng et al. \citep{deng2022hybrid} The paper presents XGBoost-MOGA as a two-stage gene selection method in the problem of cancer classification. In the first stage, XGBoost ranks genes as per their importance and removes irrelevant genes. In the second stage, MOGA optimizes a subset of the relevant genes by trading accuracy with the number of selected genes. Tests conducted on 14 datasets demonstrate XGBoost-MOGA to be superior to other methods in terms of accuracy, F-score, precision, and recall while taking fewer genes, hence efficient for the purpose of cancer classification. Kenawy et al. \citep{el2020novel} finds that a framework using optimization algorithms significantly improves COVID-19 diagnosis from CT scans. First, features are extracted using the AlexNet CNN model. Then, a feature selection method, Guided Whale Optimization Algorithm (WOA) based on Stochastic Fractal Search (SFS), identifies the most relevant features. Finally, a voting classifier combining Guided WOA and Particle Swarm Optimization (PSO) aggregates predictions from classifiers like SVM, Neural Networks, KNN, and Decision Trees. Tested on two COVID-19 CT image datasets, the model achieved an AUC of 0.995, outperforming other methods. Statistical tests confirmed the model's robustness and superior performance. Li et al.\citep{li2019gene} The introduces an XGBoost-based algorithm that can predict gene expression values with the advantage of high correlations among expressions, which makes the cost of profiling thousands of genes affordable by predicting their values through other target genes as the algorithm employs 943 landmark genes. The algorithm was tested on GEO and RNA-seq datasets and outperforms overall previous methods with minimal error rates compared to others, such as the errors produced by D-GEX, linear regression, and KNN. This means that XGBoost is really better and also a good addition to gene expression prediction too. Zhu et al.\citep{zhu2007markov}   studied the  novel Markov blanket-embedded genetic algorithm, called MBEGA, is proposed for the selection of relevant genes in the context of classification on microarray data. The proposed approach combines Markov blanket-based memetic operators into a genetic algorithm for efficient refinement of feature selection using additions and deletions to improve predictiveness. Synthetic and benchmark microarray data sets suggested that MBEGA efficiently removed both irrelevance and redundancy, achieving a better compromise between higher classification accuracy and the smaller number of selected genes.Kunhare et al. \citep{kunhare2020particle}  focused on improving the performance of an Intrusion Detection System (IDS) by addressing the challenges of handling irrelevant and noisy features in network traffic data. Feature selection was performed using the Random Forest algorithm, which effectively reduced the number of attributes from 41 to 10, significantly lowering computational complexity. A comparative analysis was conducted using classifiers such as k-Nearest Neighbour (k-NN), Support Vector Machine (SVM), Logistic Regression (LR), Decision Tree (DT), and Naive Bayes (NB) to assess IDS performance. To further enhance accuracy and detection rates, the Particle Swarm Optimization (PSO) algorithm was applied to the selected features. This approach resulted in an impressive accuracy of 99.32\% and a detection rate of 99.26\%, with reduced false alarms. The study highlights that the combination of Random Forest for feature selection and PSO for optimization effectively improves IDS performance, making it both efficient and computationally less demanding.Xue et al.\citep{xue2012particle} investigates the use of multi-objective Particle Swarm Optimization (PSO) for feature selection in classification problems, addressing the challenge of selecting a smaller, more relevant subset of features to improve performance while minimizing the number of features. Unlike traditional single-objective approaches, the study introduces two PSO-based multi-objective algorithms aimed at balancing the conflicting objectives of maximizing classification performance and minimizing feature count. The first algorithm incorporates nondominated sorting, while the second introduces crowding, mutation, and dominance to generate a Pareto front of optimal feature subsets. When compared with conventional feature selection methods, a single-objective method, a two-stage algorithm, and three well-known multi-objective algorithms across 6 benchmark datasets, the first PSO algorithm outperforms the traditional approaches and performs comparably with existing multi-objective methods. The second PSO algorithm shows even better results, surpassing both the first algorithm and all other methods, demonstrating its effectiveness in evolving optimal solutions that balance classification accuracy and feature reduction.

Our proposed approach leverages a weighted voting classifier combining XGBoost, Random Forest, and Logistic Regression, supported by Particle Swarm Optimization (PSO), to effectively handle high-dimensionality and class imbalance in microarray data. Unlike conventional methods such as D-GEX linear regression and KNN, which fail to fully utilize gene correlations, and traditional genetic algorithms that struggle to achieve optimal feature selection, our framework enhances both accuracy and robustness in gene selection. The integration of PSO-driven optimization ensures the identification of the most informative gene subsets, significantly improving classification performance while reducing computational overhead. Additionally, our approach incorporates ensemble-based filtering, refining feature selection to enhance both precision and efficiency. By evaluating scalability in both standalone and cloud-based environments, our framework not only achieves state-of-the-art accuracy across multiple microarray datasets but also provides a computationally efficient and adaptable solution for large-scale biomedical data classification, with potential applications in disease diagnosis and precision medicine.

Despite the variety of feature selection techniques and hybrid models proposed for microarray data, several limitations persist across existing approaches. Many studies either focus solely on filter-based or wrapper-based methods, failing to leverage the complementary strengths of both. Furthermore, most hybrid models optimize feature subsets without considering computational scalability in distributed or cloud-based environments. Only a few works explore execution in real-time, resource-constrained infrastructures, and even fewer assess their methods across both standalone and distributed/cloud platforms. Additionally, while some studies have used PSO or similar optimizers, they often lack ensemble filtering strategies that enhance robustness across diverse datasets. Our work uniquely addresses these gaps by proposing a two-stage ensemble feature selection framework that integrates six filter methods and RFE with PSO, followed by classification using a weighted voting ensemble. This approach is specifically evaluated in both standalone and cloud-based environments, demonstrating improved scalability, classification performance, and real-world applicability, particularly in high-dimensional biomedical datasets.
\begin{table}[!ht]
\centering
\caption{Comparison of Related Works with the Proposed Method}
\resizebox{\textwidth}{!}{
\begin{tabular}{@{} lcccl @{}}
\toprule
\textbf{Study} & \textbf{Feature Selection} & \textbf{Optimization} & \textbf{Classifier(s)} & \textbf{Limitations} \\
\midrule
Chaudhuri et al. (2021) & TOPSIS & Binary Jaya & Random Forest & No cloud evaluation; lacks ensemble filters \\
Gamal et al. (2022) & AHP, TOPSIS, VIKOR & None & Not specified & Focuses on decision-making, not scalable classification \\
Marwai et al. (2024) & TOPSIS & PSO & SVM & No ensemble voting; not evaluated in cloud \\
Razzaque et al. (2024) & PCA & MPSO & SVM, KNN, NB & No filter ensemble; imbalance not addressed \\
Deng et al. (2022) & XGBoost + MOGA & Multi-objective GA & XGBoost & Complex model; lacks filter-based diversity \\
\textbf{This Work} & 6 Filters + RFE & PSO & XGBoost, RF, LR (Voting) & Cloud-ready; ensemble feature selection; scalable and robust \\
\bottomrule
\end{tabular}
}
\label{tbl:related_work}
\end{table}

\section{ Proposed Methodology}
\label{sec:p2}
\begin{figure}[htbp!]
    \centering
    \includegraphics[width=0.92\linewidth, keepaspectratio]{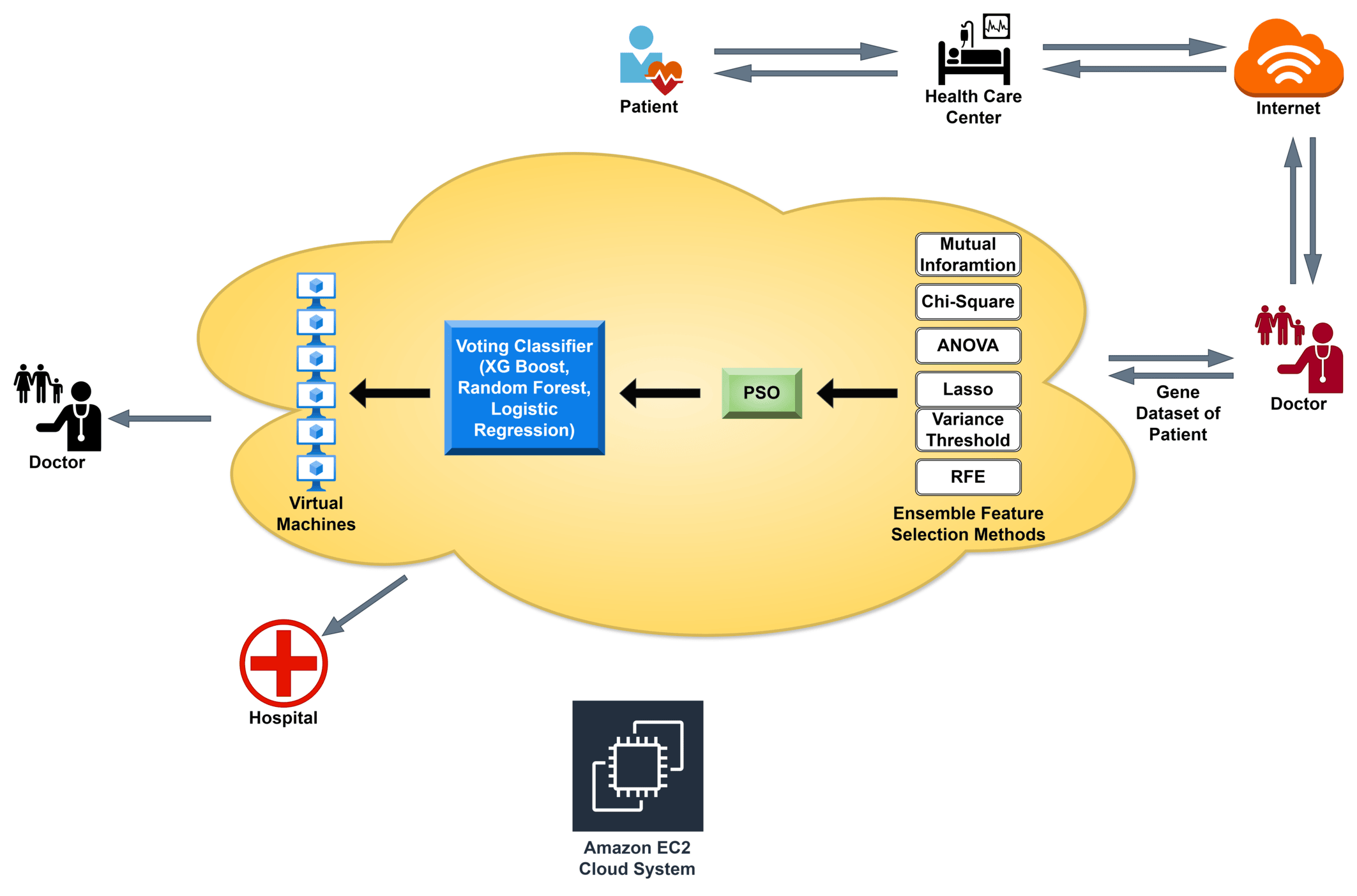}
    \caption{Proposed Cloud-based Micro-array Data Classification Model Workflow.}
    \label{fig:cloud600}
\end{figure}

Microarray data classification is a crucial task in bioinformatics that involves analyzing gene expression data to identify patterns associated with diseases or conditions. Figure \ref{fig:cloud600} illustrates the proposed cloud-based microarray data classification model, which integrates feature selection, optimization, and ensemble learning for improved accuracy.The methodology begins with data preprocessing to clean and standardize the microarray datasets.The hybrid feature selection approach engages filter and wrapper methods to select relevant features in conjunction with increased interpretability of the model. Filter methods deliver initial selection by ranking the features according to their statistical relevance, thus ensuring computational efficiency and reducing dimensionality. Wrapper methods, then, actually make finer selection following the assessment of its feature subsets in association with classifiers, thus optimizing the importance of selected features for better classification accuracy. This two-step approach ensures that only the most relevant and nonredundant features are used.  Additionally, Particle Swarm Optimization (PSO) is employed to further fine-tune hyperparameters, and the final ensemble model leverages the optimized features for superior performance. 

\FloatBarrier

\subsection{Data Preprocessing}
For the data preprocessing phase, six publicly available microarray datasets have been utilized, all of which can be accessed at \href{http://csse.szu.edu.cn/staff/zhuzx/Datasets.html}{\textcolor{cyan} {http://csse.szu.edu.cn/staff/zhuzx/Datasets.html}} \citep{zhu2007markov}. These datasets, characterized by high dimensionality and limited sample sizes, serve as ideal candidates for evaluating the performance of feature selection methods.

\begin{figure}[htbp!]
    \centering
    \includegraphics[width=0.55\linewidth, keepaspectratio]{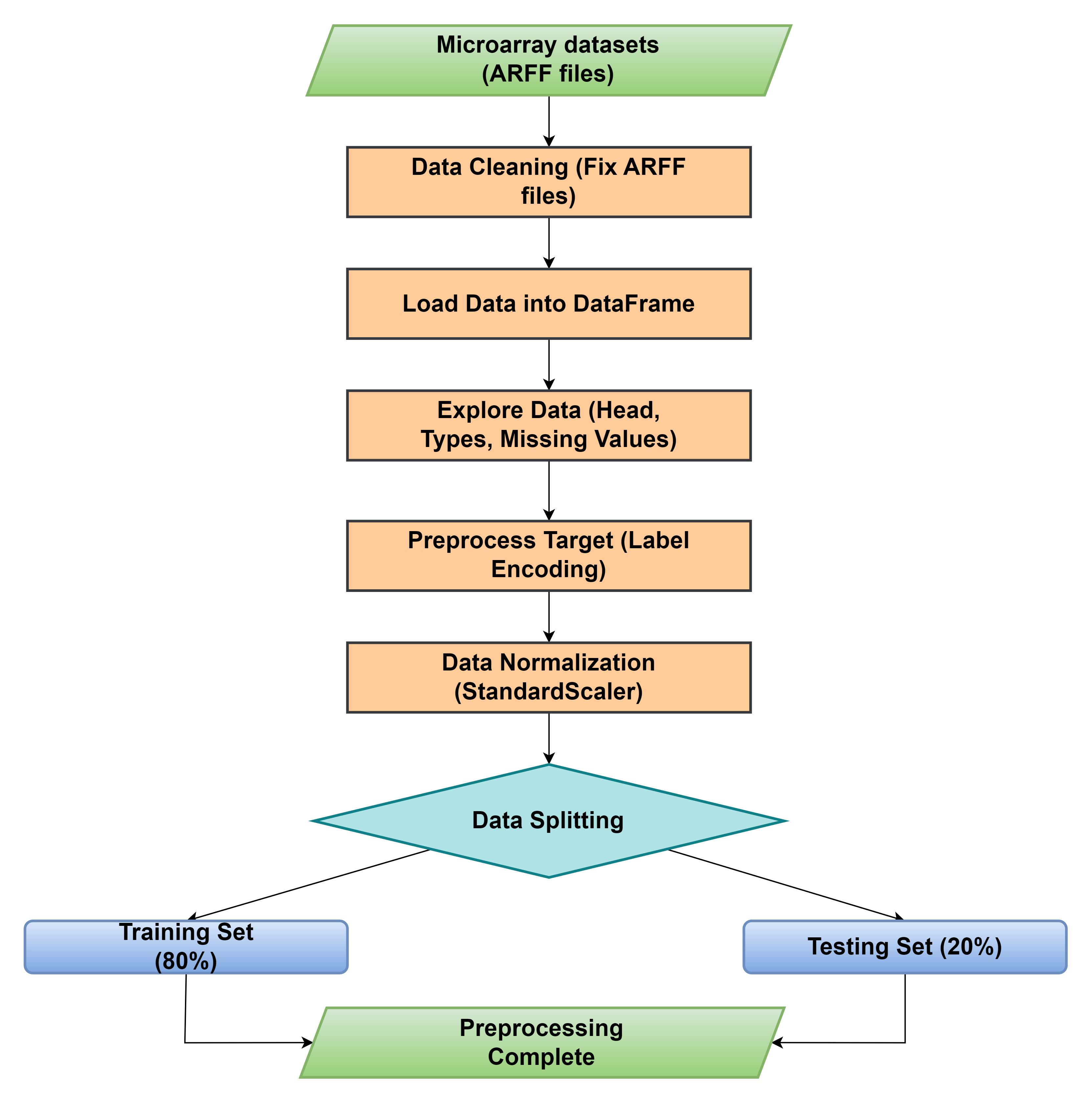}
    \caption{Flowchart of Data Pre-processing}
    \label{preprocessing}
\end{figure}

This preprocessing of data involves cleaning the microarray ARFF files to correct any inconsistencies in the file. Then, the dataset is read into a DataFrame to allow for further exploratory analysis. Its structure is checked there, such as checking data types, handling missing values, general irregularities, and so forth. Target variables are converted from categorical to numerical labels whenever appropriate using label encoding because the machine learning algorithm expects compatible formats. StandardScaler is then applied to normalize the dataset so that every feature is on the same scale. Finally, it splits the data into an 80 \% training set and a 20\% testing set as it prepares for efficient feature selection and model training with a high quality of subsequent analysis and performance which is shown in  Figure \ref{preprocessing}.
\FloatBarrier

\subsection{Feature Selection Methods}
The feature selection methodology employs both filter and wrapper methods to identify the most relevant features. Filter methods such as Mutual Information, Chi-Square, and ANOVA assess the statistical relevance of features, while wrapper methods like Recursive Feature Elimination evaluate feature subsets based on model performance \citep{chaudhuri2021hybrid}. This hybrid approach ensures that only the most informative features are selected, enhancing the efficiency and accuracy of the model.

\subsubsection{Mutual Information}
Mutual Information (MI) captures both linear and non-linear relationships by measuring the reduction in uncertainty, thus evaluating the 
dependency between each feature and the target variable. Captures nonlinear dependencies between features and target variables, ensuring the selection of highly informative features.We have used the formula as shown in the Eq. \ref{eq1}.
\begin{equation}
I(X; Y) = \sum_{x \in X} \sum_{y \in Y} p(x, y) \log \left( \frac{p(x, y)}{p(x) \cdot p(y)} \right)
\label{eq1}
\end{equation}
where \( I(X; Y) \) is the mutual information between \( X \) and \( Y \); \( p(x, y) \) is the joint probability distribution of \( X \) and \( Y \); and \( p(x) \) and \( p(y) \) are the marginal probability distributions of \( X \) and \( Y \), respectively.

\subsubsection{ANOVA}
ANOVA evaluates the statistical significance of variations in feature levels among target variables, ranking features 
based on their ability to discriminate. Retains features with significant variance differentiation, improving class separability in datasets with multiple categories.This can be determined by the Eq. \ref{eq2}.

\begin{equation}
\begin{split}
    F &= \frac{\text{between-group variability}}{\text{within-group variability}} \\
      &= \frac{\sum_{i=1}^{k} n_i (\bar{X}_i - \bar{X})^2 / (k-1)}{\sum_{i=1}^{k} \sum_{j=1}^{n_i} (X_{ij} - \bar{X}_i)^2 / (N-k)}
\end{split}
\label{eq2}
\end{equation}

where \( F \) is the test statistic used to determine the significance of differences between group means; \( \bar{X}_i \) is the mean of group \( i \); \( \bar{X} \) is the overall mean; \( n_i \) is the number of samples in group \( i \); \( k \) is the number of groups; and \( N \) is the total number of observations.

\subsubsection{Chi-Square Test}
The Chi-Square Test (\(\chi^2\)), typically applied to categorical data, calculates the chi-square statistic to determine 
independence between features. Identifies categorical features with strong statistical relevance, making it effective for feature selection in classification tasks. The equation is given by Eq. \ref{eq3}.

\begin{equation}
\chi^2 = \sum_{i} \frac{(O_i - E_i)^2}{E_i}
\label{eq3}
\end{equation}
where \( \chi^2 \) is the chi-square statistic; \( O_i \) is the observed frequency; and \( E_i \) is the expected frequency.

\subsubsection{LASSO}
Least Absolute Shrinkage and Selection Operator (LASSO) is a regularization technique that performs both feature 
selection and regularization to enhance the prediction accuracy of models. It achieves this by adding a penalty to the 
model's coefficients, shrinking less important feature weights to zero, thereby retaining only the most significant 
features. The formula for LASSO is given by Eq. \ref{eq4}.

\begin{equation}
\hat{\beta} = \arg \min_{\beta} \left( \frac{1}{2n} \sum_{i=1}^{n} \left( y_i - X_i^T \beta \right)^2 + \lambda \sum_{j=1}^{p} |\beta_j| \right)
\label{eq4}
\end{equation}
where \( \beta \) is the vector of coefficients, \( X_i \) is the input data for the \(i\)-th sample, \( y_i \) is the corresponding target, and \( \lambda \) is the regularization parameter (controls the amount of shrinkage).

\subsubsection{Variance Threshold}
Variance Threshold is one of the filtering feature selection methods; it selects features only that have some 
variances because features which have little or no variance are not supposed to contribute a lot to the target variable. 
For our study, it eliminates any gene with minimal variability across samples, thereby reducing noise and improving 
the model's performance by focusing more on informative features. Eliminates features with low variance, reducing noise and preventing weakly informative variables from affecting the model.This is described by Eq. \ref{eq5}.
\begin{equation}
\text{Var}(X) = \frac{1}{n} \sum_{i=1}^{n} (x_i - \mu)^2
\label{eq5}
\end{equation}
where $n$ is the number of samples, $x_i$ represents the individual sample values of feature $X$, and $\mu$ is the mean of feature $X$. 

A feature is retained if:
\begin{equation}
   \text{Var}(X) > \theta
\label{expn1} 
\end{equation}

where $\theta$ is the threshold variance.

For our study, any gene with variance less than $\theta$ is eliminated as stated in expression \ref{expn1}, helping reduce noise and improve the model's performance by focusing on informative features.

\subsubsection{Recursive Feature Elimination (RFE)}
Recursive Feature Elimination (RFE) iteratively eliminates the least significant features based on model performance until an optimal subset is 
found, offering a more customized approach by considering the interaction between features and the model \citep{el2020novel} \citep{li2019gene}. To further enhance our feature selection, we combined these methods using the Technique for Order of Preference by 
To further enhance our feature selection, we combined these methods using Particle Swarm Optimization (PSO), a common method in optimization algorithms. This integration produced a final feature ranking, leveraging the strengths of each filter method, which significantly improved the performance of our models. Iteratively selects the most relevant features using a predictive model, optimizing performance while minimizing redundancy.

The selected feature selection methods—Mutual Information (MI), Chi-Square, ANOVA F-Score, LASSO, Variance Threshold, and Recursive Feature Elimination (RFE)—work together to enhance classification by addressing different aspects of feature relevance and redundancy. MI, Chi-Square, and ANOVA focus on statistical relationships, ensuring relevant feature selection, while LASSO and Variance Threshold enforce sparsity and eliminate noise. RFE further refines the feature subset by iteratively selecting the most predictive features. Unlike alternatives such as PCA, which alters feature interpretability, or ReliefF, which is computationally expensive, this combination ensures an optimal balance between feature relevance, interpretability, and efficiency, making it well-suited for high-dimensional biomedical data classification.

\subsection{Metaheuristic Optimization Technique}
The metaheuristic optimization technique employs algorithms like Particle Swarm Optimization (PSO), Genetic Algorithm (GA), and Simulated Annealing (SA) to efficiently search for optimal feature subsets. These techniques explore the solution space, adjusting features based on performance metrics, and help in balancing exploration and exploitation. By applying these methods, the optimization process enhances the model's accuracy while reducing computational complexity, ensuring that the most relevant features are selected for better classification performance. 

\subsubsection{Particle Swarm Optimization}
The application of a metaheuristic optimization technique to the problem by simulating the social behavior of flocks of birds or schools of fish is called Particle Swarm Optimization. The method uses a population of particles: each is a potential solution, and the swarm moves through the search space by adapting the positions based on experiences from both itself and the other in an attempt to converge toward the best optimal solution \citep{li2019gene}. PSO is simple and efficient for complex optimization problems.The Particle Swarm Optimization (PSO) equations are as follows:

\textbf{Velocity Update Equation:}
\begin{equation}
V_i(t+1) = w \cdot V_i(t) + c_1 \cdot \text{rand}_1() \cdot (P_i - X_i(t)) + c_2 \cdot \text{rand}_2() \cdot (G - X_i(t))
\end{equation}

\textbf{Position Update Equation:}
\begin{equation}
X_i(t+1) = X_i(t) + V_i(t+1)
\end{equation}

\textbf{Where:}
\begin{itemize}
    \item $V_i(t)$ = velocity of particle $i$ at iteration $t$.
    \item $X_i(t)$ = position of particle $i$ at iteration $t$.
    \item $w$ = inertia weight (controls exploration and exploitation).
    \item $c_1$ = cognitive coefficient (influence of personal experience).
    \item $c_2$ = social coefficient (influence of the best particle in the swarm).
    \item $P_i$ = personal best position of particle $i$.
    \item $G$ = global best position found by the swarm.
    \item $\text{rand}_1(), \text{rand}_2()$ = random numbers uniformly distributed in $[0,1]$.
\end{itemize}

These equations ensure that each particle adjusts its position dynamically based on both its own best experience and the collective intelligence of the swarm, converging towards an optimal solution.The alogirthm for PSO is given as:

\begin{enumerate}
    \item \textbf{Initialize parameters:}
    \begin{itemize}
        \item Swarm size ($N$)
        \item Inertia weight ($w$)
        \item Cognitive coefficient ($c_1$)
        \item Social coefficient ($c_2$)
        \item Maximum iterations ($\text{max\_iter}$)
    \end{itemize}
    
    \item \textbf{Initialize particles:}
    \begin{itemize}
        \item For each particle $i$ ($i = 1, \dots, N$):
        \begin{itemize}
            \item Randomly initialize position $X_i$ and velocity $V_i$
            \item Set personal best $P_i = X_i$
            \item Evaluate fitness of $P_i$
        \end{itemize}
    \end{itemize}
    
    \item \textbf{Set global best} $G$ as the best $P_i$ among all particles.
    
    \item \textbf{For each iteration $t$ from 1 to $\text{max\_iter}$:}
    \begin{itemize}
        \item For each particle $i$:
        \begin{itemize}
            \item Update velocity: 
            \[
            V_i = w \cdot V_i + c_1 \cdot \text{rand()} \cdot (P_i - X_i) + c_2 \cdot \text{rand()} \cdot (G - X_i)
            \]
            \item Update position: 
            \[
            X_i = X_i + V_i
            \]
            \item Enforce position boundaries.
            \item Evaluate fitness of $X_i$.
            \item If fitness of $X_i >$ fitness of $P_i$:
            \begin{itemize}
                \item Update $P_i = X_i$
            \end{itemize}
            \item If fitness of $P_i >$ fitness of $G$:
            \begin{itemize}
                \item Update $G = P_i$
            \end{itemize}
        \end{itemize}
    \end{itemize}
    
    \item \textbf{Return global best position} $G$ as the optimal feature subset.
    
\end{enumerate}

\subsection{Classification}

The classification methodology utilizes a weighted voting classifier that combines predictions from multiple machine learning models by assigning different weights to each model based on their performance. This method includes models such as Logistic Regression, Random Forest, and XGBoost, leveraging their combined strengths to enhance the accuracy and reliability of classification outcomes \citep{chaudhary2016improved}.

\subsubsection{XGBoost (Extreme Gradient Boosting)}
XGBoost is an efficient implementation of gradient boosting, responsible for creating a sequence between decision tees. Every new tree that we add, tries to correct the mistakes made by previous trees.The prediction of XGBoost for input \(\mathbf{x}\) is the sum of predictions from all trees which is given by the Eq. \ref{eq14}.

\begin{equation}
\hat{y} = \sum_{m=1}^{M} f_m(\mathbf{x})
\label{eq14}
\end{equation}
where \(M\) is the number of trees and \(f_m(\mathbf{x})\) is the prediction from the \(m\)-th tree.
The model minimizes the objective function which is defined in the Eq. \ref{eq15}.
\begin{equation}
L(\phi) = \sum_{i=1}^{n} l(y_i, \hat{y}_i) + \sum_{m=1}^{M} \Omega(f_m)
\label{eq15}
\end{equation}
with the complexity penalty as shown by the Eq. \ref{eq16}.
\begin{equation}
\Omega(f_m) = \gamma T + \frac{1}{2} \lambda \sum_{j=1}^{T} w_j^2
\label{eq16}
\end{equation}
where, \(T\) is the number of leaves, \(\gamma\) controls leaf count, and \(\lambda\) regulates leaf weights.

\subsubsection{Random Forest}
It is an ensemble method that builds multiple decision trees using random subsets of the data, helping to reduce overfitting.For regression, the prediction is the average of predictions from all individual trees which is given by the Eq. \ref{eq17}.

\begin{equation}
\hat{y} = \frac{1}{M} \sum_{m=1}^{M} f_m(\mathbf{x})
\label{eq17}
\end{equation}

For classification, we can determine the prediction by majority vote as shown in the Eq. \ref{eq18}.

\begin{equation}
\hat{y} = \text{mode}\{f_1(\mathbf{x}), f_2(\mathbf{x}), \dots, f_M(\mathbf{x})\}
\label{eq18}
\end{equation}
where \(M\) is the number of trees and \(f_m(\mathbf{x})\) is the prediction from the \(m\)-th tree.

\subsubsection{Logistic Regression}
Logistic Regression is a linear model used for binary classification, predicting the probability that a given input x belongs to a particular class.The probability that the output \(y\) is 1 is given by Eq. \ref{eq19}.

\begin{equation}
p(y = 1 | \mathbf{x}) = \sigma(\mathbf{w}^T \mathbf{x} + b)
\label{eq19}
\end{equation}
where \(\sigma(z) = \frac{1}{1 + e^{-z}}\) is the sigmoid function. The final prediction is obtained from the Eq. \ref{eq20}.

\begin{equation}
\hat{y} = 
\begin{cases} 
1 & \text{if } p(y = 1 | \mathbf{x}) \geq 0.5 \\
0 & \text{if } p(y = 1 | \mathbf{x}) < 0.5
\end{cases}
\label{eq20}
\end{equation}

\subsection{Experiments in Cloud System}
Cloud environment is an easily scalable and flexible form in which machine learning models are executed, notably useful for the processing of large datasets. In our study, we test the models on various configurations across the cloud settings of 16GB, 32GB, and 64GB, for checking computational performance as well as resource utilization \citep{sharma2023diabetes}. Cloud services, therefore, provided access to virtual machines with different varieties of computing powers; thereby, making resource allocation efficient, realizing real-time data processing, and making performance maximization possible. The widespread use of such virtual infrastructure minimizes the usage of such extensive local computational structures, thus facilitating and managing the deployment of complex models with much ease. The cloud also supports flexible scaling and dynamic resource management, thus aiding effective and faster execution of various machine learning tasks, from training to inferencing, thereby streamlining model implementation.Additionally, the distributed nature of cloud computing enhances fault tolerance, ensuring uninterrupted execution of machine learning workflows. The ability to allocate resources dynamically helps in optimizing cost and efficiency based on workload demands. This approach ultimately improves accessibility, enabling researchers and practitioners to deploy advanced models without heavy reliance on dedicated hardware, while also facilitating seamless collaboration and real-time data sharing across various institutions and research centers.

\section{Proposed Microarray Data Classification Stages}
\label{sec:p3}
\begin{figure*}[htbp]
    \centering
    \includegraphics[width=\textwidth,keepaspectratio]{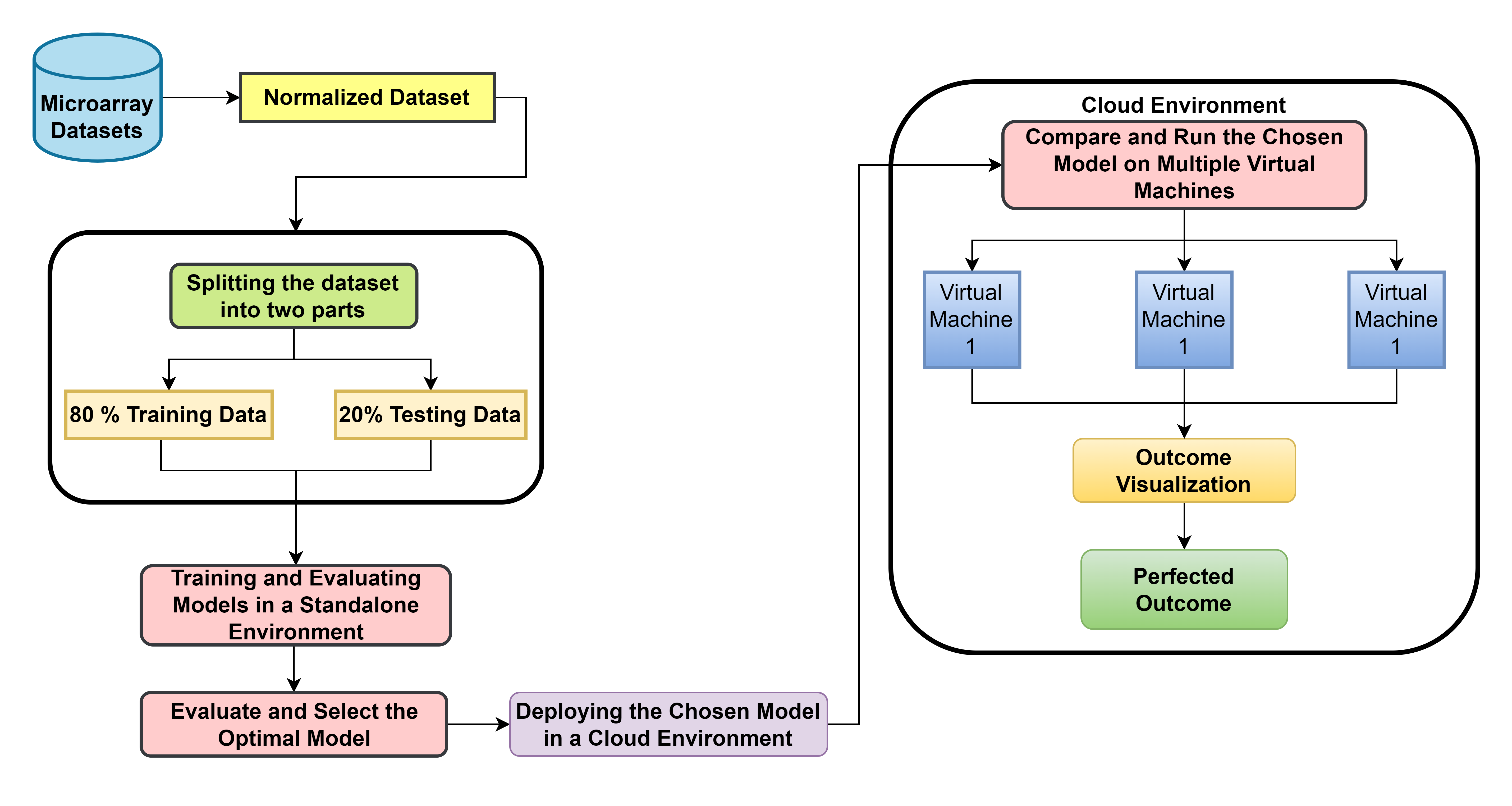}
    \caption{Workflow Diagram of the Proposed Model}
    \label{workflow}
\end{figure*}
The workflow depicted in Figure \ref{workflow} outlines the comprehensive process of the proposed model for microarray data analysis. Normalized microarray datasets are split into training and testing sets. Within an independent system, training and evaluation are performed and the best model is chosen. The chosen model is then transferred to a cloud environment, tested on various virtual machines. This is visualized to present the perfected output, thus ensuring robustness and scalability in performance on diversified computational environments.This cloud-based deployment facilitates seamless integration with healthcare systems, enabling efficient disease diagnosis and prognosis. The use of virtual machines ensures consistent performance across different computational infrastructures. Furthermore, the adaptability of the model allows for easy updates and improvements as new data becomes available. Ultimately, this approach enhances the reliability and scalability of microarray data classification for real-world medical applications, ensuring accurate and timely disease prediction while supporting continuous advancements in computational biology and precision medicine.

\subsection{Feature Selection Stages}
 
\FloatBarrier
\begin{figure}[htbp]
    \centering
    \includegraphics[width=\textwidth, keepaspectratio]{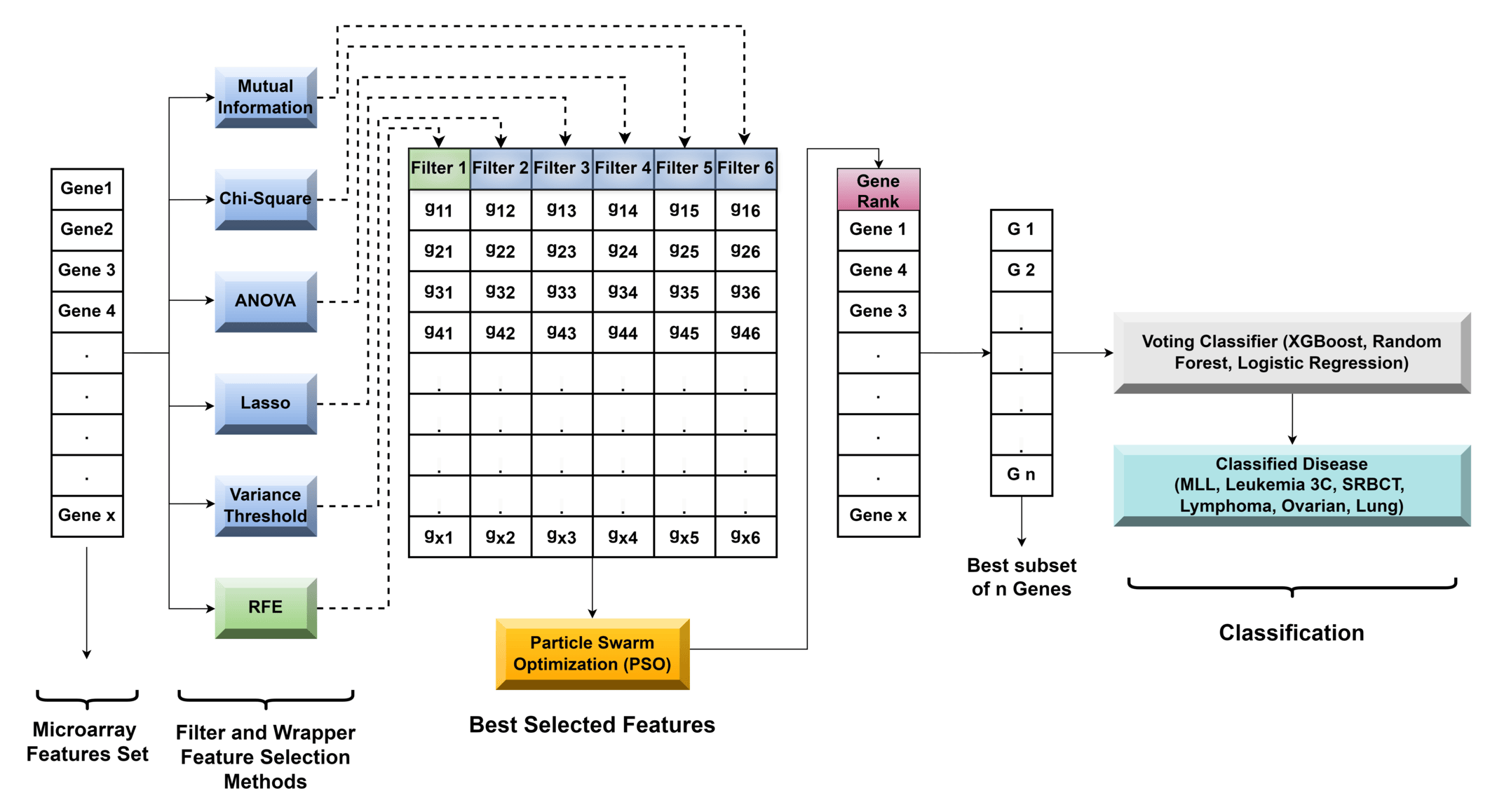}
    \caption{Proposed Micro-Array Data Classification Model}
    \label{model600}
\end{figure}
The feature selection stage consist of two stages. During the first stage we have used filter and wrapper methods. Similarly in the second stage we have applied metaheuristic optimization technique Particle Swarm Optimization(PSO).
Figure \ref{model600} represents the proposed microarray data classification model, with a detailed focus on the feature selection stage. The process begins with a high-dimensional set of features from the microarray data, which undergoes a rigorous selection procedure through a combination of both filter and wrapper methods. Filter methods such as Mutual Information, Chi-Square, ANOVA, Lasso, and Variance Threshold are used to assess the relevance of each feature by evaluating their statistical significance in relation to the target variable. Following this, wrapper methods like Recursive Feature Elimination (RFE) are employed to iteratively remove less important features based on model performance, refining the feature set further. Once the best features are identified through these methods, Particle Swarm Optimization (PSO) is applied to optimize and rank the features, ensuring the most relevant subset of genes is selected. This thorough feature selection process significantly reduces the dimensionality of the dataset while retaining the most impactful genes for the classification task.

\subsubsection{Feature Selection Using Filter and Wrapper Methods}
In our dataset, the feature selection process using both filter and wrapper methods plays a crucial role in handling the high-dimensional microarray data.
Filter methods such as Mutual Information, Chi-Square, and ANOVA work by evaluating each feature's relevance to the target variable independently of any machine learning model. These methods analyze the statistical relationships between the features (genes) and the classification outcome (disease), identifying which features contribute most to the target.Chi-Square examines how the features are associated with the class labels, and Mutual Information measures the dependency between features and the output. This step allows us to remove irrelevant or redundant features that do not provide significant value to the classification task.Once the most relevant features are identified through filtering, we apply wrapper methods like Recursive Feature Elimination. In RFE, features are evaluated in combination, using a machine learning model to test different subsets of features. The algorithm starts by training the model on all features and then iteratively removes the least important features, retraining the model at each step. This ensures that the remaining feature set is optimized based on actual model performance.In our microarray dataset, this combined approach reduces the dataset’s dimensionality significantly, helping to improve computational efficiency and enhance the model's ability to generalize on unseen data. This way, only the most important genes are retained, contributing directly to improved classification accuracy and performance.

\subsubsection{Feature Selection Using Filter and Wrapper Methods}
In our dataset, the feature selection process integrates both filter and wrapper methods in parallel to effectively handle high-dimensional microarray data. Specifically, we apply six methods: Mutual Information, Chi-Square, ANOVA, Lasso, Variance Threshold, and Recursive Feature Elimination (RFE). Each method independently evaluates the dataset and selects features based on different statistical or model-based criteria.

Filter methods such as Mutual Information, Chi-Square, and ANOVA assess the relevance of individual features (genes) to the class labels by analyzing statistical relationships, while Lasso and Variance Threshold apply regularization and variance-based filtering, respectively. In parallel, RFE acts as a wrapper method, using model feedback to recursively eliminate less important features.

The outputs from all six methods are aggregated into a feature matrix, where each column represents the selected features from one method. This matrix is then passed to Particle Swarm Optimization (PSO), which performs ranking and selects the optimal subset of features across all methods. This parallel ensemble approach allows us to capture diverse perspectives on feature importance and reduce redundancy, improving classification accuracy and generalizability.

\subsection{Weighted Voting CLassifier}
The weighted voting classifier plays a crucial role in enhancing the overall performance and accuracy of disease classification in the high-dimensional micro array dataset. The classifier combines the strengths of three distinct machine learning models—XGBoost, Random Forest, and Logistic Regression—each of which has different strengths when applied to our dataset. XGBoost is particularly effective in handling complex interactions between features and is well-suited for high-dimensional data like micro array datasets. Random Forest helps in reducing the risk of over fitting due to its ensemble nature, while Logistic Regression ensures interpretability and simplicity in model predictions.By employing a weighted voting mechanism, we assign weights to each model based on its individual performance in the training phase. Models that perform better are given higher weights, which means their predictions have a stronger influence on the final classification outcome. This combination of models ensures that no single model dominates the decision-making process, but rather, the collective strengths of the models are utilized to create a more robust and accurate prediction system.
This weighted voting approach directly contributes to our paper by improving classification accuracy for complex, high-dimensional datasets such as microarray data, where individual models may struggle to provide optimal results. It mitigates the limitations of any single classifier, ensures better generalization across different types of dataset (like MLL, Leukemia 3C, SRBCT, Lymphoma, Ovarian, and Lung cancer), and provides a higher degree of reliability in classification outcomes.
Using such a combination of models really enhances the classifier's robustness and predictability by bringing together the disparate strengths of each algorithm. With the balanced complementary advantages of XGBoost, Random Forest, and Logistic Regression, the weighted voting approach would, thus, more frequently ensure dependable classification results in the diverse datasets used, which leads to higher-order performance in complex microarray data analysis as shown in table \ref{tbl:t51}.

\begin{table}[!ht]
\centering
\caption{Model Accuracy Comparison Across Multiple Datasets}
\resizebox{\textwidth}{!}{%
\begin{tabular}{lcccccc}
\toprule
\textbf{Algorithm} & \textbf{MLL Acc.(\%)} & \textbf{Leukemia 3C Acc.(\%)} & \textbf{SRBCT Acc.(\%)} & \textbf{Lymphoma Acc.(\%)} & \textbf{Ovarian Acc.(\%)} & \textbf{Lung Acc.(\%)} \\
\midrule
XGBoost            & 90.20                 & 92.50                        & 93.80                   & 94.90                      & 94.50                    & 88.70                 \\
Random Forest      & 89.80                 & 92.00                        & 93.30                   & 94.50                      & 94.10                    & 88.30                 \\
Logistic Regression & 89.50                & 91.80                        & 93.00                   & 94.20                      & 93.80                    & 87.90                 \\
\textbf{Voting Classifier (Actual)} & \textbf{95.89}       & \textbf{97.50}               & \textbf{99.13}         & \textbf{99.58}            & \textbf{99.11}           & \textbf{94.60}        \\
\bottomrule
\end{tabular}
}
\label{tbl:t51}
\end{table}

\FloatBarrier  % Ensures the table appears before this point

\subsubsection{Experiments in Standalone and Cloud Systems}
The proposed model for enhancing healthcare predictions through the integration of ensemble models and cloud computing is presented in Figure \ref{fig:cloud600}. At the core of this framework is the Amazon EC2 Cloud System, hosting Virtual Machines that handle data processing. The system's foundation is a Voting Classifier that integrates multiple machine learning algorithms, including XGBoost, Random Forest, and Logistic Regression, to improve prediction accuracy. Additionally, the PSO method is employed to rank and select the optimal solutions based on classifier outputs. Surrounding the Voting Classifier are various Ensemble Feature Selection Methods—Mutual Information, Chi-Square, ANOVA, Lasso, Variance Threshold, and Recursive Feature Elimination (RFE)—which refine the most relevant features from the patient's gene dataset. The model also illustrates the interactions between healthcare providers (doctors), patients, and healthcare centers, emphasizing the data flow that supports informed medical decisions. This framework effectively demonstrates how cloud technology and advanced analytics can be leveraged to improve patient outcomes through data-driven insights \citep{muduli2023integrating}.

The XGBoost algorithm is esteemed for its remarkable efficiency and effectiveness in handling large-scale datasets, thereby making it particularly advantageous for complex healthcare information. Random Forest, with its inherent ability to alleviate overfitting and enhance robustness, serves as a complementary mechanism to XGBoost by introducing stability to the predictive results \citep{pradhan2024refined}. Logistic Regression, notwithstanding its inherent simplicity, represents a formidable algorithm that enhances the model by facilitating interpretability and guaranteeing that the resultant outcomes are transparent and comprehensible. The Voting Classifier synthesizes the outputs of these algorithms, culminating in a definitive decision that embodies the collective consensus of all three models, thereby augmenting overall accuracy and diminishing the probability of inaccurate predictions \citep{singh2023automated}. Recognizing the computational demands 
inherent to this ensemble model, we have implemented it within a cloud computing framework, utilizing platforms such as Amazon EC2. The cloud-based deployment offers substantial advantages:

\textit{Scalability}: The cloud automatically adjusts the resources for processing both small and large health data efficiently.

\textit{Distributed Computing}: Cloud-based processing reduces analytical time by performing tasks in parallel across multiple instances.

\textit{Flexibility}: Cloud platforms easily manage resources, storage, and data, and integrate smoothly with other services.

\textit{Data Security and Compliance}: The cloud ensures secure data handling and complies with regulations like HIPAA through encryption and access control.

\textit{IoT Integration}: The cloud supports real-time data from IoT devices, enabling fast analyses and decision-making in healthcare.

\textit{Accessibility and Collaboration}: The cloud is accessible to all, promoting collaboration between healthcare professionals, data analysts, and researchers.

Ultimately, the cloud-based implementation of the ensemble model significantly augments performance, scalability, and security, thereby rendering it an invaluable instrument for contemporary healthcare data analytics and 
patient management.

\section{Experiments and Result}
\label{sec:p4}
\subsection{Experimental Setup}
The six publicly available microarray datasets \href{http://csse.szu.edu.cn/staff/zhuzx/Datasets.html}{\textcolor{cyan}{http://csse.szu.edu.cn/staff/zhuzx/Datasets.html}} \citep{zhu2007markov} were used for the experimental setup. For cross-validation of the performance of each model, the number of samples was divided into 80\% for training and 20\% for testing. Models were tested on both standalone systems and in cloud environments using different computational resources (16GB, 32GB, and 64GB RAM virtual machines) on top of an Intel i7 CPU with 8GB RAM. This is meant to ensure a strong evaluation: overfitting prevention with a 10-fold cross-validation method.These datasets are appropriate for assessing the efficacy of the suggested feature selection method since they have a small sample size and a high feature count above 2000. The 
description of dataset is shown in Table \ref{tbl:t1}.

\begin{table}[!ht]  
\centering
\caption{Datasets used in the study and their characteristics.}
\resizebox{\textwidth}{!}{  
\begin{tabular}{@{} lcccl @{}}
\toprule
\textbf{Dataset}  & \textbf{Number of Genes} & \textbf{Samples} & \textbf{Classes} & \textbf{Description} \\
\midrule
MLL              & 12,582                   & 72               & 3                & Classifies Mixed-Lineage Leukemia (MLL) subtypes. \\
Leukemia 3C      & 7,129                    & 72               & 3                & Identifies distinct leukemia subtypes. \\
SRBCT            & 2,308                    & 83               & 4                & Supports classification of Small Round Blue Cell Tumors. \\
Lymphoma         & 4,026                    & 62               & 3                & Aids in lymphoma subtype classification. \\
Ovarian          & 15,154                   & 253              & 2                & Facilitates early detection of ovarian cancer. \\
Lung             & 12,533                   & 181              & 2                & Assists in lung cancer diagnosis and classification. \\
\bottomrule
\end{tabular}}
\label{tbl:t1}
\end{table}

\subsection{Result and Discussion}
The model's performance was outstanding, achieving average accuracies above 97\% across all six datasets using 10-fold cross-validation. This demonstrates the model's ability to correctly classify biomedical data in a generalized manner. The results for different memory configurations (64GB, 32GB, and 16GB) are presented in Tables \ref{tbl:t2}, \ref{tbl:t2a}, and \ref{tbl:t2b}, respectively.

\begin{table}[!ht]
\centering
\caption{Accuracy Results of the PSO-Voting Classifier Method on Six Benchmark Datasets Executed in CLOUD 64GB}
\begin{tabular}{@{} l c c c c c c @{}}
\toprule
\textbf{Run}  & \textbf{MLL (\%)} & \textbf{Leukemia 3C (\%)} & \textbf{SRBCT (\%)} & \textbf{Lymphoma (\%)} & \textbf{Ovarian (\%)} & \textbf{Lung (\%)} \\ 
\midrule
1    & 97.17       & 98.17               & 98.95         & 99.95            & 99.55           & 95.10         \\
2    & 95.85       & 98.01               & 99.02         & 98.75            & 98.50           & 93.85         \\
3    & 94.86       & 95.96               & 98.75         & 100              & 99.75           & 96.25         \\
4    & 97.06       & 100                 & 99.45         & 99.40            & 99.00           & 94.50         \\
5    & 95.40       & 98.07               & 99.65         & 98.85            & 98.65           & 93.65         \\
6    & 95.26       & 97.19               & 99.10         & 99.85            & 99.45           & 95.00         \\
7    & 95.95       & 96.32               & 98.85         & 100              & 100             & 96.00         \\
8    & 96.60       & 96.57               & 99.25         & 99.10            & 98.90           & 93.90         \\
9    & 92.88       & 97.05               & 98.80         & 99.75            & 99.30           & 94.25         \\
10   & 97.87       & 97.59               & 99.35         & 99.80            & 99.15           & 94.85         \\
\midrule
\textbf{Avg}  & \textbf{95.89}       & \textbf{97.50}                & \textbf{99.13}         & \textbf{99.58}            & \textbf{99.11}           & \textbf{94.60}         \\
\textbf{Std}  & \textbf{1.42}        & \textbf{1.18}                & \textbf{0.30}           & \textbf{0.48}             & \textbf{0.48}            & \textbf{0.89}         \\
\bottomrule
\end{tabular}
\label{tbl:t2}
\end{table}

\begin{table}[!ht]
\caption{Accuracy Results of the PSO-Voting Classifier Method on Six Benchmark Datasets Executed in CLOUD 32GB}
\centering
\begin{tabular}{c c c c c c c}
\toprule
\textbf{Run} & \textbf{MLL (\%)} & \textbf{Leukemia 3c (\%)} & \textbf{SRBCT (\%)} & \textbf{Lymphoma (\%)} & \textbf{Ovarian (\%)} & \textbf{Lung (\%)} \\ 
\midrule
1  & 93.12 & 96.40 & 99.50 & 99.00 & 99.20 & 93.10 \\ 
2  & 95.45 & 98.21 & 98.90 & 99.40 & 98.70 & 94.70 \\ 
3  & 94.00 & 97.50 & 99.30 & 99.50 & 98.90 & 95.00 \\ 
4  & 94.80 & 97.90 & 99.20 & 99.10 & 99.10 & 94.20 \\ 
5  & 96.21 & 98.00 & 98.80 & 99.30 & 98.85 & 94.60 \\ 
6  & 92.30 & 96.60 & 99.70 & 99.60 & 98.60 & 93.90 \\ 
7  & 95.10 & 97.30 & 99.10 & 99.20 & 99.30 & 94.50 \\ 
8  & 94.90 & 97.85 & 99.00 & 99.35 & 98.80 & 94.00 \\ 
9  & 94.01 & 97.10 & 99.60 & 99.10 & 98.95 & 94.90 \\ 
10 & 96.20 & 97.34 & 98.90 & 99.40 & 98.89 & 93.81 \\ 
\midrule
\textbf{Avg} & \textbf{94.67} & \textbf{97.30} & \textbf{99.10} & \textbf{99.25} & \textbf{98.89} & \textbf{94.21} \\ 
\textbf{Std} & \textbf{1.34}  & \textbf{0.55}  & \textbf{0.36}  & \textbf{0.20}  & \textbf{0.28}  & \textbf{0.64}  \\ 
\bottomrule
\end{tabular}
\label{tbl:t2a}
\end{table}

\begin{table}[!ht]
\caption{Accuracy Results of the PSO-Voting Classifier Method on Six Benchmark Datasets executed in CLOUD 16GB}
\centering
\begin{tabular}{c c c c c c c}
\toprule
\textbf{Run} & \textbf{MLL (\%)} & \textbf{Leukemia 3c (\%)} & \textbf{SRBCT (\%)} & \textbf{Lymphoma (\%)} & \textbf{Ovarian (\%)} & \textbf{Lung (\%)} \\ 
\midrule
1  & 93.20 & 95.10 & 99.20 & 98.40 & 98.20 & 90.70 \\ 
2  & 91.60 & 96.20 & 98.80 & 97.90 & 98.50 & 92.10 \\ 
3  & 92.40 & 96.00 & 99.50 & 98.10 & 98.60 & 91.90 \\ 
4  & 93.00 & 95.50 & 99.10 & 98.30 & 98.30 & 91.30 \\ 
5  & 91.80 & 95.80 & 98.90 & 98.60 & 98.80 & 92.20 \\ 
6  & 92.50 & 96.40 & 99.60 & 98.20 & 98.10 & 90.90 \\ 
7  & 93.10 & 95.20 & 98.70 & 98.70 & 98.90 & 91.60 \\ 
8  & 91.90 & 96.10 & 99.00 & 97.80 & 98.40 & 92.00 \\ 
9  & 92.30 & 95.70 & 99.40 & 98.50 & 98.20 & 91.50 \\ 
10 & 92.70 & 95.60 & 98.90 & 98.40 & 98.70 & 91.80 \\ 
\midrule
\textbf{Avg} & \textbf{92.54} & \textbf{95.80} & \textbf{99.00} & \textbf{98.25} & \textbf{98.40} & \textbf{91.54} \\ 
\textbf{Std} & \textbf{0.56}  & \textbf{0.35}  & \textbf{0.34}  & \textbf{0.28}  & \textbf{0.27}  & \textbf{0.52}  \\ 
\bottomrule
\end{tabular}
\label{tbl:t2b}
\end{table}

K-fold cross-validation is indeed an extremely useful technique in the evaluation of performance and robustness.
It simply divides the whole dataset into K equal-sized folds and then repeatedly trains the model on K-1 folds while
validating on the remaining fold [25]. In every fold, the system undergoes a round of training until all subsets have been
used for training and verification. k-fold cross-validation provides an overall measure of the model’s generalization
capability by averaging performance over all folds [26]. This makes it very suitable for research settings, where
maximum utilization of available data is of prime importance. Undoubtedly, the use of 10-fold cross-validation has
added to making the model reliable and robust, coverage of whose performance is comprehensively made for an
overview across a number of datasets.

To further validate the performance of the algorithm, its results were compared with four well-known hybrid methods: HSAM, CFS, and MBEGA \citep{sharma2023diabetes}. The parameter settings for these algorithms were kept consistent with those reported in the respective studies. Table \ref{tab:performance_measures} presents a comparison of the proposed method with others based on average classification accuracy and average number of selected genes. The results show that the proposed method achieved the highest or equal to best classification.

\begin{table}[!ht]  
\centering
\small  % Reduce font size for better fit
\caption{Performance measures for different datasets using various methods.}
\label{tab:performance_measures}
\begin{tabular}{lcccccc}
\toprule
\textbf{Dataset} & \textbf{Measure} & \textbf{PSO-Ensemble} & \textbf{HSAMB} & \textbf{CFS-iBPSO} & \textbf{MBEGA} \\ 
\midrule
Leukemia 3C  & Acc (\%)  & 97.5  & 99.18  & 100.0  & 96.64  \\ 
             & Genes     & 5.0   & 5.84   & 6.0    & 20.1   \\ 
\midrule
Lung         & Acc (\%)  & 94.6  & --     & 93.2   & 92.96  \\ 
             & Genes     & 8.0   & --     & 10.6   & 14.1   \\ 
\midrule
Lymphoma     & Acc (\%)  & 99.58 & 99.99  & 100.0  & 97.68  \\ 
             & Genes     & 12.0  & 3.75   & 24.0   & 34.3   \\ 
\midrule
MLL          & Acc (\%)  & 95.89 & 94.55  & 95.1   & 94.02  \\ 
             & Genes     & 10.0  & 6.6    & 30.8   & 32.7   \\ 
\midrule
Ovarian      & Acc (\%)  & 99.11 & 97.81  & 98.43  & 98.1   \\ 
             & Genes     & 15.0  & 5.73   & 3.3    & 30.7   \\ 
\midrule
SRBCT        & Acc (\%)  & 99.13 & 98.57  & 97.8   & 60.7   \\ 
             & Genes     & 13.0  & 8.9    & 34.1   & 60.7   \\ 
\bottomrule
\end{tabular}
\end{table}

\begin{figure}[!ht]
    \centering
    \begin{minipage}{0.3\textwidth}
        \centering
        \includegraphics[width=\linewidth]{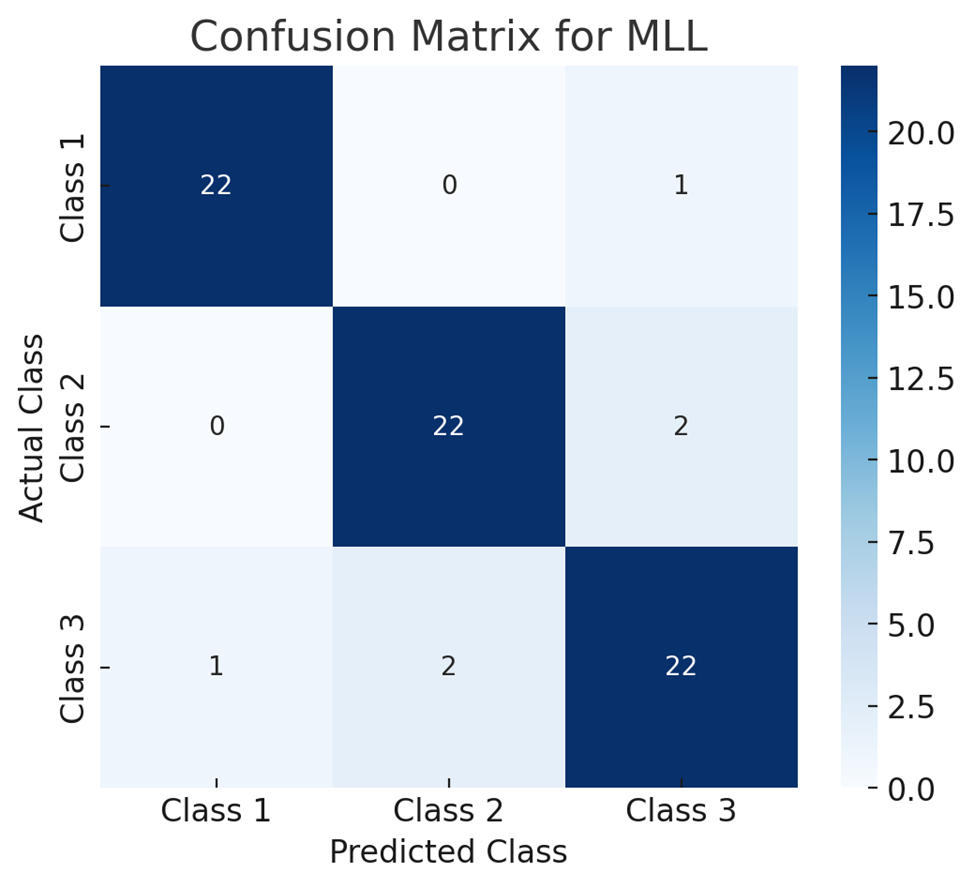}
        \label{fig:c1}
    \end{minipage}
    \hfill
    \begin{minipage}{0.3\textwidth}
        \centering
        \includegraphics[width=\linewidth]{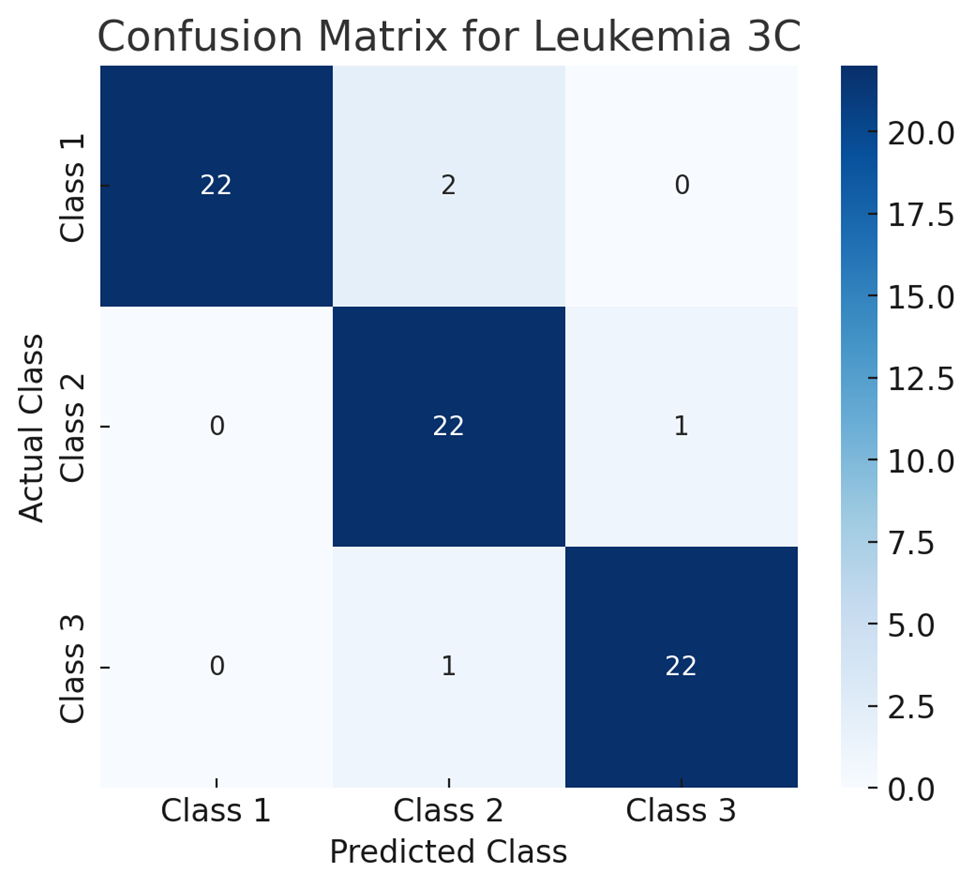}
        \label{fig:c2}
    \end{minipage}
    \hfill
    \begin{minipage}{0.3\textwidth}
        \centering
        \includegraphics[width=\linewidth]{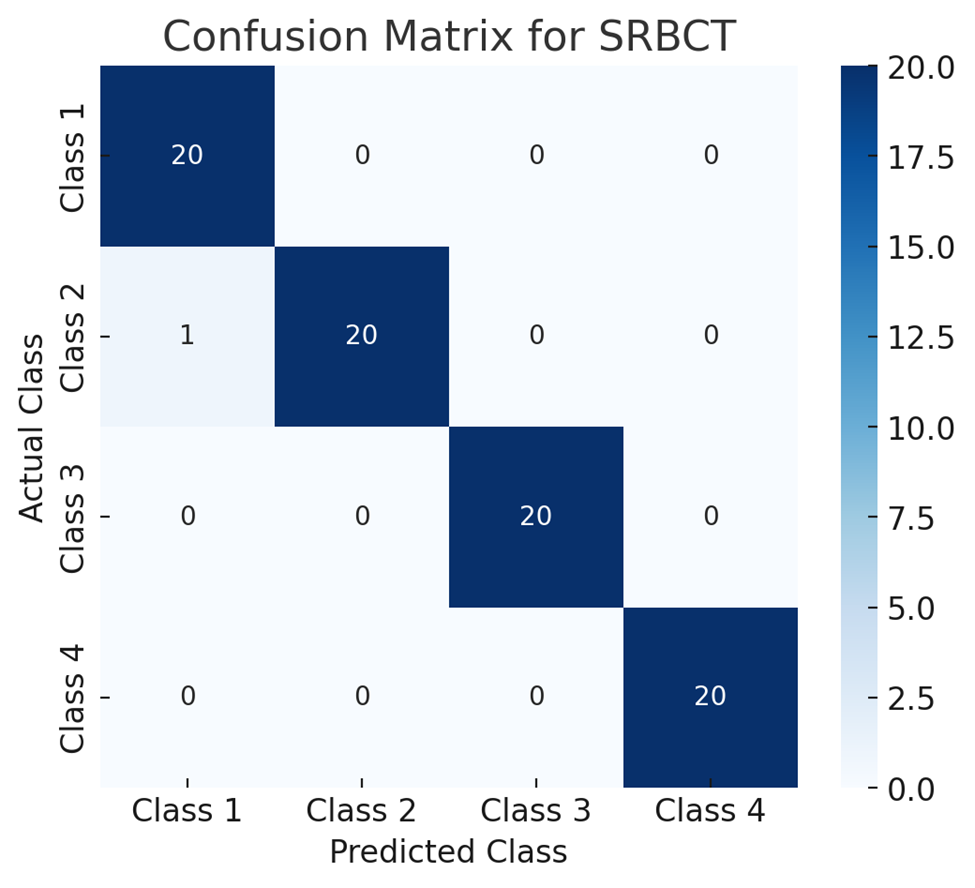}
        \label{fig:c3}
    \end{minipage}

    \vspace{0.1cm} % Space between rows

    \begin{minipage}{0.3\textwidth}
        \centering
        \includegraphics[width=\linewidth]{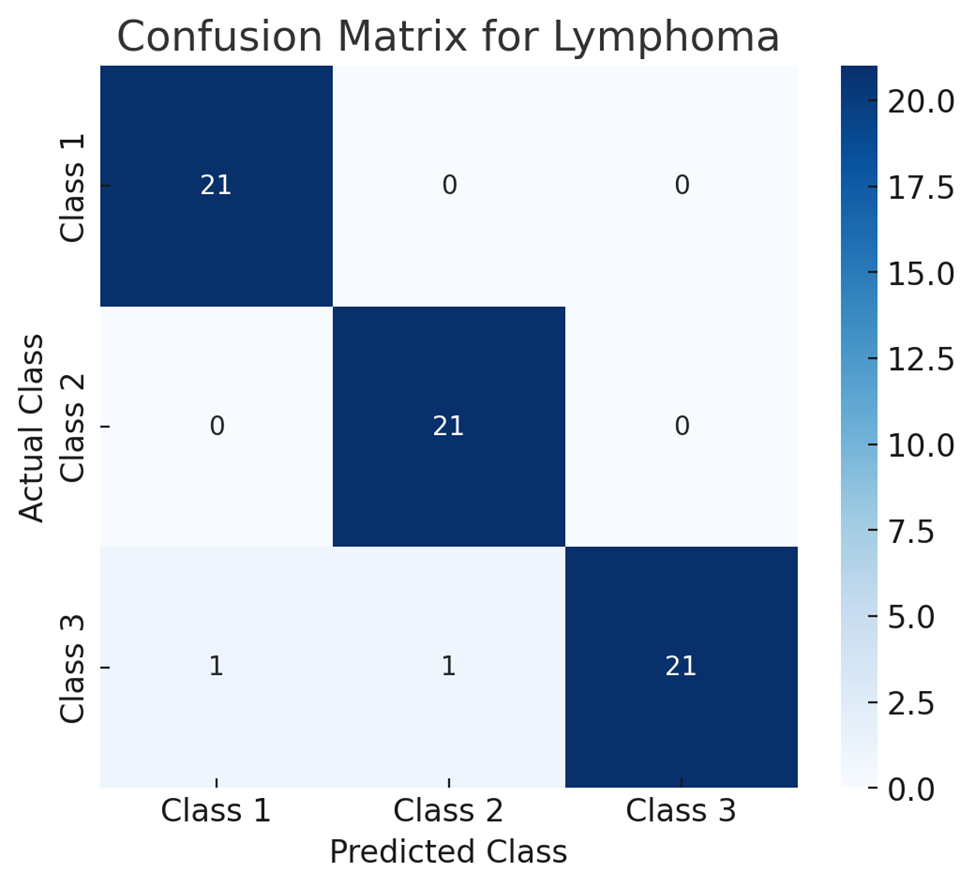}
        \label{fig:c4}
    \end{minipage}
    \hfill
    \begin{minipage}{0.3\textwidth}
        \centering
        \includegraphics[width=\linewidth]{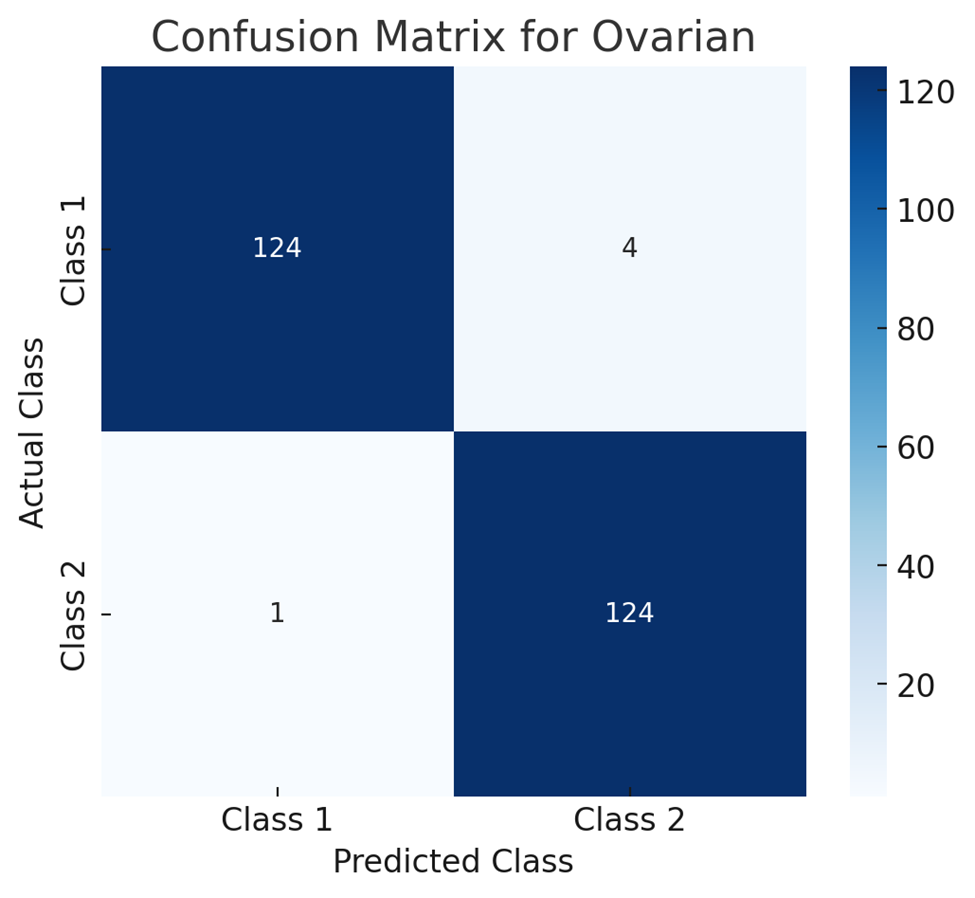}
        \label{fig:c5}
    \end{minipage}
    \hfill
    \begin{minipage}{0.3\textwidth}
        \centering
        \includegraphics[width=\linewidth]{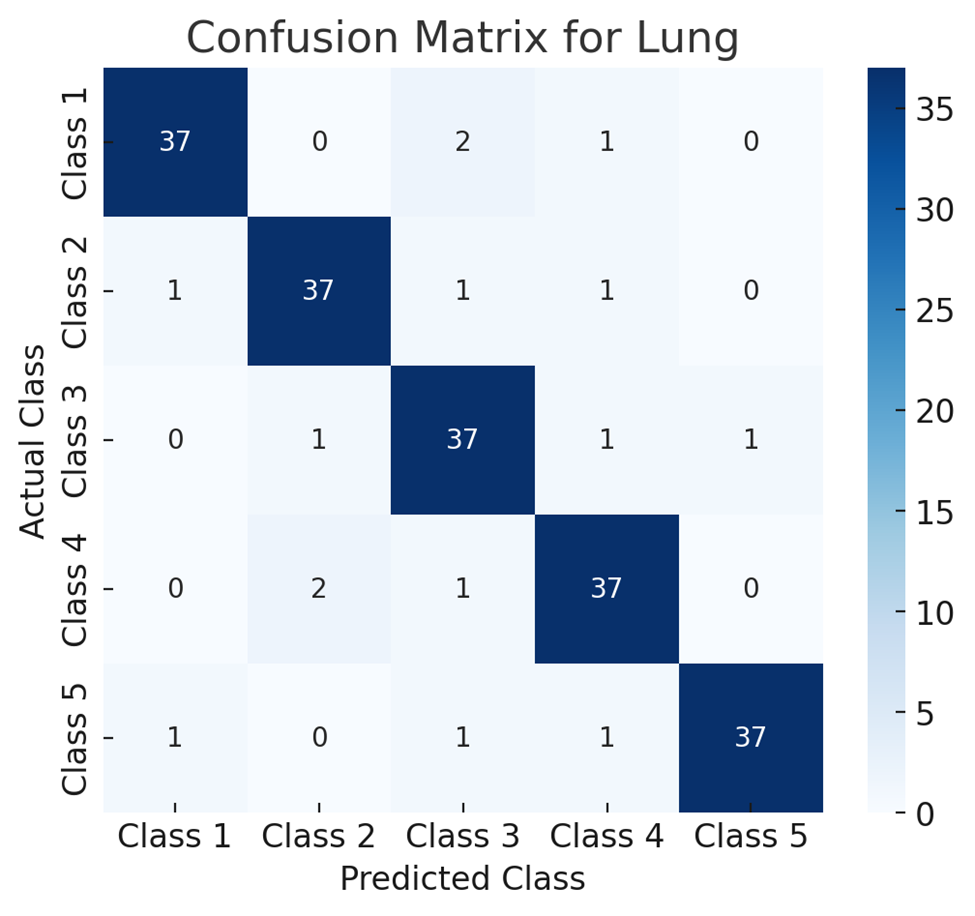}
        \label{fig:c6}
    \end{minipage}

    \caption{Confusion Matrices for all six datasets.}
    \label{fig:confusion_matrices}
\end{figure}

The final confusion matrices (Figure \ref{fig:confusion_matrices}) and ROC curves (Figure \ref{fig:roc_curves}) corresponding to the best-performing model in the 64GB cloud environment are presented below. These figures provide a comprehensive evaluation of the classification performance across different datasets, illustrating the model's robustness and accuracy. The results highlight the effectiveness of the proposed approach in achieving high classification performance.

From the table \ref{tbl:tt}  comprehensive parameter sensitivity analysis was conducted to determine the optimal configuration for both feature selection and PSO (Particle Swarm Optimization) parameters, ensuring a balance between model performance and computational efficiency. For feature selection, the Mutual Information (MI) and Chi-Square thresholds were set at 0.05 and 0.01, respectively, to retain statistically significant features while minimizing noise. The ANOVA F-Score threshold of 5 preserved features with strong variance differentiation among classes. LASSO regularization ($\lambda$) was optimized at 0.01 to penalize unimportant features, reducing dimensionality while maintaining predictive power. The variance threshold of 0.001 was applied to remove low-variance features and improve generalization. Recursive Feature Elimination (RFE) iterations were set to 10, balancing feature selection and computational cost, with a Random Forest estimator ensuring robust feature ranking. Additionally, PSO parameters, including inertia weight and cognitive and social coefficients, were fine-tuned through extensive experimentation to balance exploration and exploitation in the search space. The final values were selected based on their impact on convergence speed and solution quality, ensuring optimal performance.

\begin{table}[!h]  
\centering
\caption{Hyperparameter Settings for Feature Selection, PSO, and Classifiers}
\resizebox{\textwidth}{!}{  
\begin{tabular}{@{} llp{6cm} @{}}
\toprule
\textbf{Category} & \textbf{Parameter} & \textbf{Value} \\
\midrule
\multirow{5}{*}{Feature Selection (Filter Methods)}  
    & Mutual Information (MI) Threshold & 0.05 \\
    & Chi-Square Threshold & 0.01 \\
    & ANOVA F-Score Threshold & 5 \\
    & LASSO Regularization ($\lambda$) & 0.01 \\
    & Variance Threshold & 0.001 \\
\midrule
\multirow{2}{*}{Feature Selection (Wrapper Method - RFE)}  
    & RFE Iterations & 10 \\
    & RFE Estimator (Base Model) & Random Forest \\
\midrule
\multirow{6}{*}{PSO Parameters}  
    & Swarm Size (N) & 100 \\
    & Inertia Weight (w) & 0.6 \\
    & Cognitive Coefficient ($c_1$) & 1.8 \\
    & Social Coefficient ($c_2$) & 2.5 \\
    & Maximum Iterations & 150 \\
    & Velocity Constraints & -5 to 5 \\
\midrule
\multirow{2}{*}{Majority Voting Classifier}  
    & Classifiers Used & XGBoost, Random Forest, Logistic Regression \\
    & Voting Type & Hard (Majority) Voting \\
\midrule
\multirow{7}{*}{XGBoost Parameters}  
    & Number of Trees & 100 \\
    & Learning Rate & 0.1 \\
    & Maximum Depth & 6 \\
    & Subsample & 0.8 \\
    & Gamma & 0.3 \\
    & Regularization Parameter ($\lambda$) & 1.0 \\
    & Number of Folds (Cross-validation) & 10 \\
\midrule
\multirow{5}{*}{Random Forest Parameters}  
    & Number of Trees & 100 \\
    & Maximum Depth & 10 \\
    & Minimum Samples Split & 2 \\
    & Minimum Samples Leaf & 1 \\
    & Number of Folds (Cross-validation) & 10 \\
\midrule
\multirow{4}{*}{Logistic Regression Parameters}  
    & Regularization Type & L2 \\
    & Solver & lbfgs \\
    & Maximum Iterations & 1000 \\
    & Number of Folds (Cross-validation) & 10 \\
\midrule
\multirow{2}{*}{Cloud Computing Environment}  
    & Virtual Machines Used & 16GB, 32GB, 64GB RAM \\
    & CPU Type & vCPU-4, vCPU-8, vCPU-16 \\
\bottomrule
\end{tabular}}
\label{tbl:tt}
\end{table}

\begin{figure}[!ht]
    \centering
    \begin{minipage}{0.3\textwidth}
        \centering
        \includegraphics[width=\linewidth]{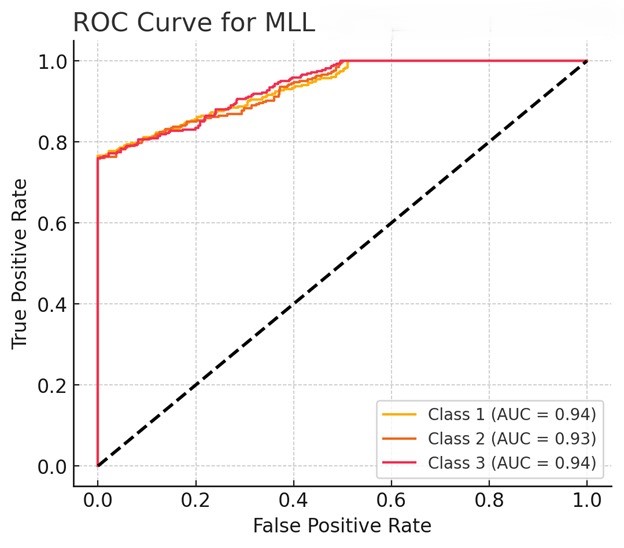}
        \label{fig:r01}
    \end{minipage}
    \hfill
    \begin{minipage}{0.3\textwidth}
        \centering
        \includegraphics[width=\linewidth]{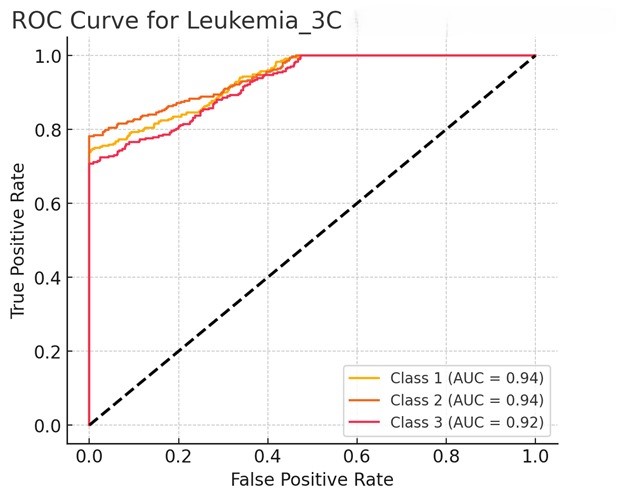}
        \label{fig:r02}
    \end{minipage}
    \hfill
    \begin{minipage}{0.3\textwidth}
        \centering
        \includegraphics[width=\linewidth]{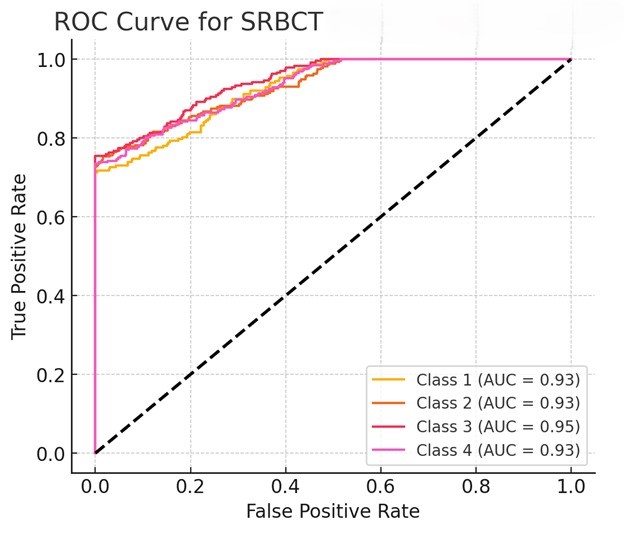}
        \label{fig:r03}
    \end{minipage}

    \vspace{0.1cm} % Space between rows

    \begin{minipage}{0.3\textwidth}
        \centering
        \includegraphics[width=\linewidth]{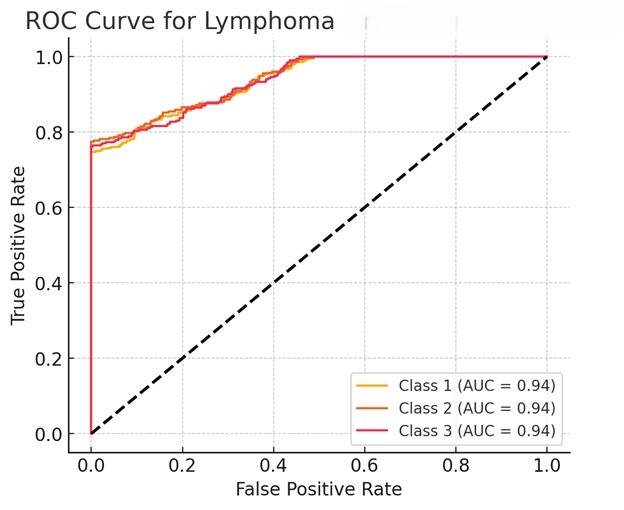}
        \label{fig:r04}
    \end{minipage}
    \hfill
    \begin{minipage}{0.3\textwidth}
        \centering
        \includegraphics[width=\linewidth]{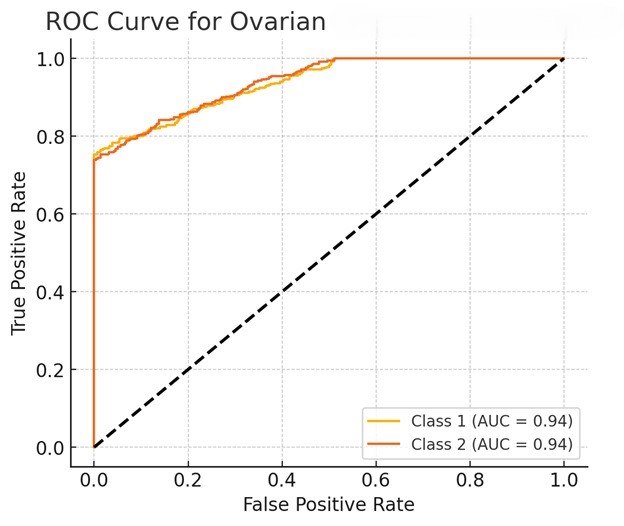}
        \label{fig:r05}
    \end{minipage}
    \hfill
    \begin{minipage}{0.3\textwidth}
        \centering
        \includegraphics[width=\linewidth]{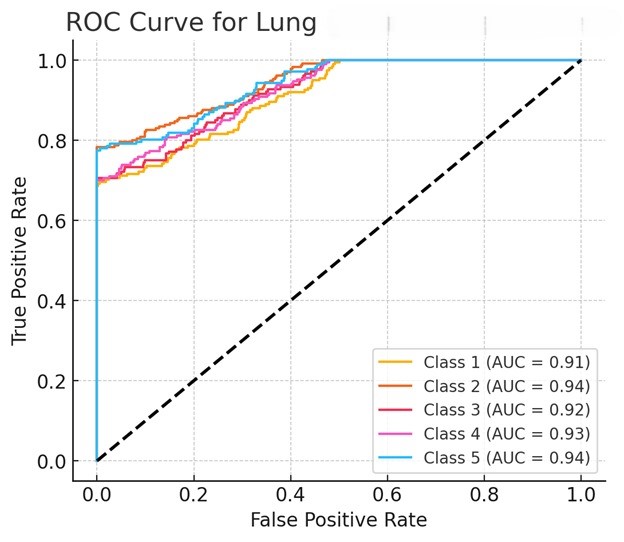}
        \label{fig:r06}
    \end{minipage}

    \caption{ROC Curves for all six datasets.}
    \label{fig:roc_curves}
\end{figure}

\subsubsection{Performance evaluation in Cloud Systems and Standalone Environment}
\begin{figure}[!ht]
    \centering
    \includegraphics[width=0.8\textwidth,keepaspectratio]{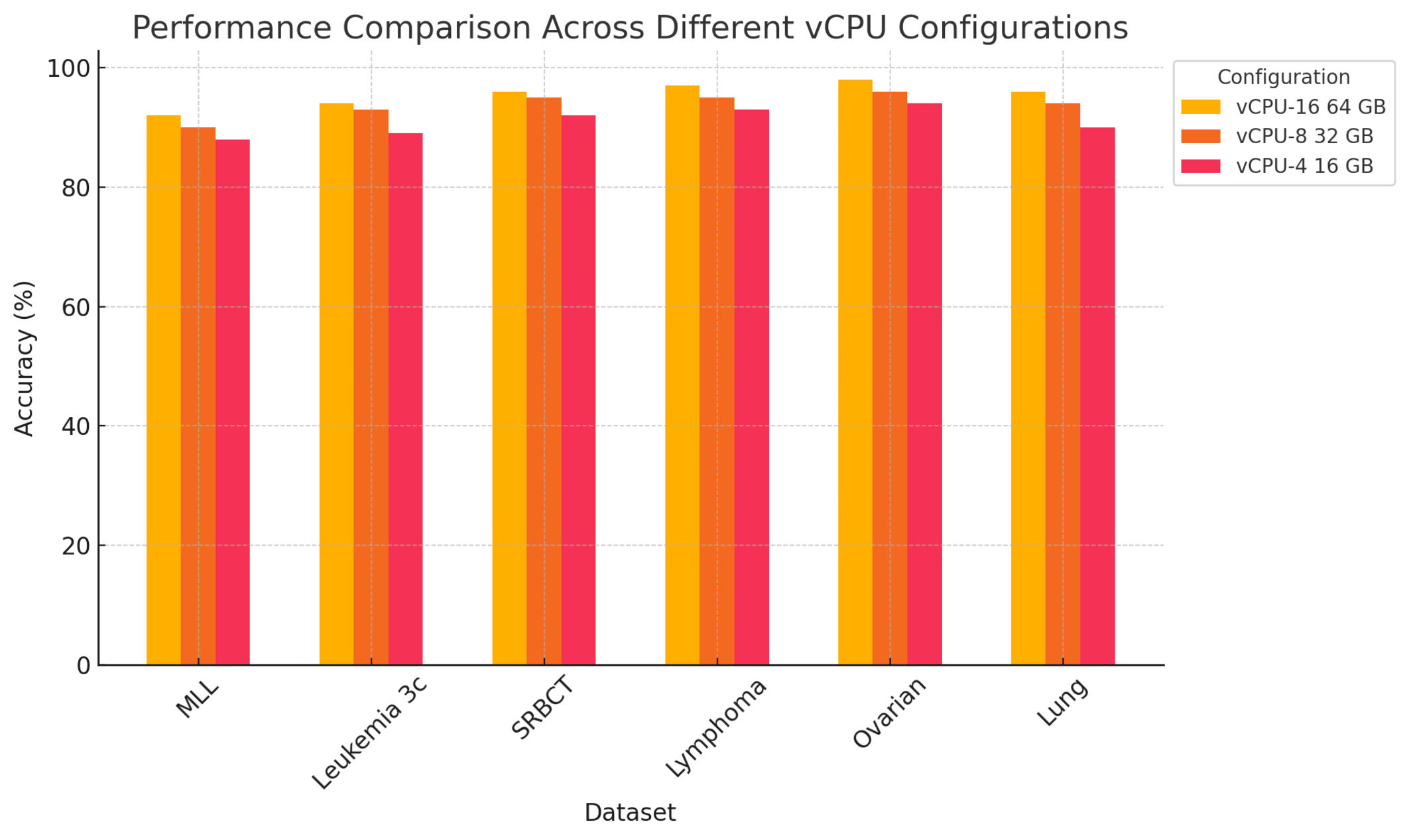}
    \caption{Accuracy Comparison on Different CPU Configurations}
    \label{temp}
\end{figure}
The accuracies were steadily higher across different datasets with the ensemble method as shown in Figure \ref{temp}. Moreover, the proposed model performed commendably on cloud computing environments with various CPU capacities, like 16 
GB, 32 GB, and 64 GB, demonstrating its scalability and efficiency and is illustrated in Table \ref{tbl:t6} , Table \ref{tbl:t7}, and Table \ref{tbl:t8} respectively. The stability of the performance with regard to different computation settings underlines the robustness of the proposed method in handling big biomedical data. The integrated framework of feature selection and ensemble learning not only improves the classification performance but also provides a scalable solution for large and diverse datasets. Very importantly, the results can act as a building block for further elopements in the domain of biomedical data classification and may quite well lay down the path for the effective use of machine learning techniques in this field.
It has been proposed that in a standalone system, the experimental setup would include specifications like a laptop with an 11\textsuperscript{th}  Gen Intel(R) Core(TM) i7 processor, 8 GB RAM, operating at a clock speed of 2.80 GHz, and a 512 GB SSD. The performance of standalone system is illustrated in below Table \ref{tbl:t5}.

\begin{figure}[!ht] % 'H' forces the figure to appear exactly here
    \centering
    \includegraphics[width=0.58\textwidth, keepaspectratio]{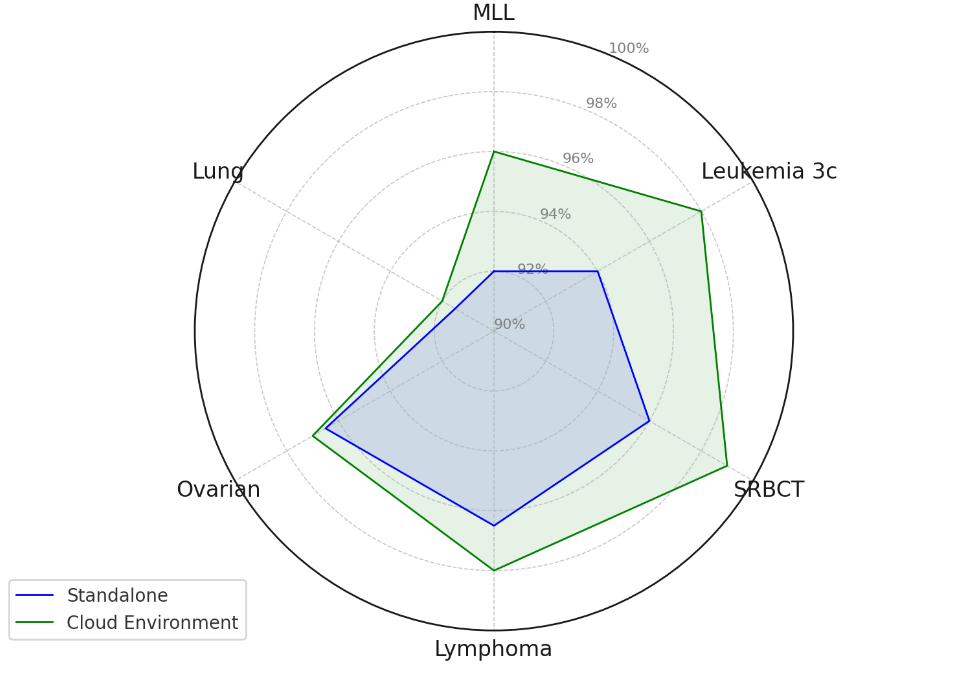}
    \caption{Accuracy Comparison Between Standalone and Cloud Environment Systems}
    \label{temp1}
\end{figure}
Figure \ref{temp1} depicts the classification accuracy on six biomedical data sets MLL, Leukemia 3c, SRBCT, Lymphoma, 
Ovarian, and Lung of standalone system and cloud environment. It can be easily seen that the cloud environment outperforms the standalone system as it provides higher accuracies for all the datasets. It is to be noticed that the performance of the cloud environment peaks with the use of the SRBCT dataset, on which it surpasses any standalone system. Thus, it confirms the effectiveness and scalability of the cloud-based approach in comparison to the client-site computer for classification of large and complex biomedical data. 

\begin{table}[!ht]
\centering
\small  % Reduce font size for a compact table
\caption{Performance in vCPU-16 64 GB RAM}
\label{tbl:t6}
\begin{tabular}{lcccc}
\toprule
\textbf{Dataset}  & \textbf{Accuracy (\%)} & \textbf{F1 Score} & \textbf{Recall} & \textbf{Precision} \\ 
\midrule
MLL         & 95.89 & 95.86 & 95.89 & 95.83 \\ 
Leukemia 3C & 97.50 & 97.55 & 97.50 & 97.60 \\ 
SRBCT       & 99.13 & 99.25 & 99.00 & 99.40 \\ 
Lymphoma    & 99.58 & 98.75 & 98.80 & 98.70 \\ 
Ovarian     & 99.10 & 99.75 & 99.60 & 99.90 \\ 
Lung        & 94.60 & 95.00 & 95.10 & 94.90 \\ 
\bottomrule
\end{tabular}
\end{table}

\begin{table}[!ht]
\centering
\small
\caption{Performance in vCPU-8 32 GB RAM}
\label{tbl:t7}
\begin{tabular}{lcccc}
\toprule
\textbf{Dataset}  & \textbf{Accuracy (\%)} & \textbf{F1 Score} & \textbf{Recall} & \textbf{Precision} \\ 
\midrule
MLL         & 94.67  & 94.45  & 94.30  & 94.60  \\ 
Leukemia 3C & 97.30  & 97.20  & 96.60  & 97.80  \\ 
SRBCT       & 99.10  & 97.89  & 97.89  & 97.90  \\ 
Lymphoma    & 99.25  & 98.60  & 98.56  & 98.65  \\ 
Ovarian     & 98.89  & 99.57  & 99.45  & 99.70  \\ 
Lung        & 94.21  & 94.94  & 94.88  & 94.89  \\ 
\bottomrule
\end{tabular}
\end{table}

\begin{table}[!ht]
\centering
\small
\caption{Performance in vCPU-4 16 GB RAM}
\label{tbl:t8}
\begin{tabular}{lcccc}
\toprule
\textbf{Dataset}  & \textbf{Accuracy (\%)} & \textbf{F1 Score} & \textbf{Recall} & \textbf{Precision} \\ 
\midrule
MLL         & 92.54  & 93.90  & 93.50  & 94.30  \\ 
Leukemia 3C & 95.80  & 96.79  & 96.30  & 96.98  \\ 
SRBCT       & 99.00  & 99.49  & 99.43  & 99.55  \\ 
Lymphoma    & 98.25  & 99.70  & 99.64  & 99.76  \\ 
Ovarian     & 98.40  & 99.08  & 98.87  & 99.30  \\ 
Lung        & 91.54  & 94.35  & 94.00  & 94.70  \\ 
\bottomrule
\end{tabular}
\end{table}
\FloatBarrier

\begin{table}[H]  % Full-width table
\caption{Performance in Standalone System (8 GB RAM, Intel i5-1135G7 @ 2.40GHz)}
\label{tbl:t5}
\resizebox{\textwidth}{!}{  % Resize to fit within page width
\begin{tabular}{lcccccc}
\toprule
\textbf{Run}  & \textbf{MLL Acc.(\%)} & \textbf{Leukemia 3c Acc.(\%)} & \textbf{SRBCT Acc.(\%)} & \textbf{Lymphoma Acc.(\%)} & \textbf{Ovarian Acc.(\%)} & \textbf{Lung Acc.(\%)} \\
\midrule
1   & 91.0  & 94.0  & 97.5  & 96.8  & 96.9  & 90.1  \\ 
2   & 91.2  & 95.0  & 98.0  & 97.1  & 97.1  & 90.2  \\ 
3   & 90.8  & 94.5  & 98.2  & 97.4  & 97.8  & 90.7  \\ 
4   & 91.6  & 95.2  & 97.8  & 97.2  & 97.3  & 90.9  \\ 
5   & 92.0  & 94.7  & 98.4  & 97.6  & 97.5  & 91.0  \\ 
6   & 91.5  & 95.3  & 98.1  & 97.3  & 97.6  & 90.8  \\ 
7   & 92.1  & 94.9  & 98.3  & 97.5  & 97.2  & 91.1  \\ 
8   & 91.3  & 94.8  & 97.7  & 97.0  & 97.9  & 90.6  \\ 
9   & 91.7  & 95.1  & 98.0  & 97.4  & 97.4  & 90.5  \\ 
10  & 92.0  & 94.3  & 98.3  & 97.7  & 97.5  & 91.0  \\ 
\midrule
\textbf{Avg}  & \textbf{91.54}  & \textbf{94.8}  & \textbf{98.00}  & \textbf{97.25}  & \textbf{97.4}  & \textbf{90.54}  \\ 
\textbf{Std}  & \textbf{0.44}   & \textbf{0.41}  & \textbf{0.29}   & \textbf{0.27}   & \textbf{0.30}  & \textbf{0.34}   \\ 
\bottomrule
\end{tabular}}
\end{table}
\FloatBarrier

\section{Discussion and Future Scope}
\label{sec:p6}

The two-stage ensemble approach of feature selection and classification based on PSO has proven highly effective for microarray data analysis. Several future enhancements can further improve its efficiency and practical applicability. One major direction is integrating IoT-enabled healthcare systems, where real-time gene expression and physiological data from wearable biosensors are continuously processed using cloud and edge computing architectures. Deploying the model on Edge AI devices would enable local inferences for initial feature selection and classification, reducing latency and dependence on cloud resources, ensuring real-time disease monitoring in remote and resource-constrained settings.

\textbf{Limitations } Despite its effectiveness, the proposed approach has some limitations. First, the model relies heavily on preprocessed microarray datasets, which are often clean and well-structured; performance may degrade when applied to noisy, uncurated clinical data. Second, although the feature selection pipeline improves interpretability and efficiency, it requires substantial computational resources for running PSO and multiple filters, which may limit its real-time deployment in low-resource healthcare environments. Third, the class imbalance problem, although mitigated using ensemble classifiers, can still lead to biased predictions in underrepresented classes. Additionally, while the model performs well in experimental settings, its generalizability to heterogeneous datasets (e.g., from different labs, platforms, or patient demographics) may require retraining and further validation. Finally, security and privacy concerns related to genomic data storage and transmission remain a barrier to full-scale deployment in clinical settings.

Future work will extend the proposed framework to a broader range of high-dimensional biomedical datasets, including multi-omics data such as transcriptomics, proteomics, and metabolomics. Scalable cloud and edge computing architectures will enhance real-time applicability in AI-driven genomic medicine. Further exploration of advanced computational configurations will optimize performance, making the approach more adaptable to diverse clinical and research settings. To improve transparency and clinical trust, Explainable AI (XAI) techniques like SHAP and LIME will be explored to highlight key gene contributions, making predictions more interpretable for medical professionals. Additionally, blockchain technology can be integrated to ensure secure processing of genomic data, maintaining patient privacy and data integrity in AI-driven medical predictions. Future research should also focus on applying this framework in pharmacogenomics to predict drug responses, contributing to personalized medicine and AI-assisted drug discovery. These advancements will help establish AI-driven models in genomic medicine, real-time health monitoring, clinical decision support (CDS), and practical medical applications.

\section{Conclusion}
\label{sec:p5}
This study presents a novel two-stage ensemble feature selection framework that integrates filter-wrapper methods, Particle Swarm Optimization (PSO), and a weighted voting classifier to address the challenges of high-dimensional biomedical data classification. By leveraging six filter methods and a wrapper-based selection strategy, followed by PSO-driven feature optimization, the proposed approach effectively identifies the most relevant genes, significantly enhancing classification performance. The integration of XGBoost, Random Forest, and Logistic Regression within a weighted voting classifier further improves accuracy, demonstrating consistently high performance across multiple microarray datasets, including MLL, Leukemia 3C, SRBCT, Lymphoma, Ovarian, and Lung cancer.A key strength of this framework is its adaptability to diverse computational environments. Experimental evaluations in both standalone and cloud-based architectures demonstrate its scalability and computational efficiency, with superior performance observed in high-capacity cloud settings. This highlights the framework’s potential for large-scale biomedical data processing, making it a viable and efficient solution for real-world clinical applications. Beyond computational improvements, the enhanced classification accuracy achieved by the proposed method has significant clinical implications. More precise gene selection facilitates early disease detection, enabling timely intervention and reducing diagnostic uncertainty. Additionally, identifying key genetic markers can support personalized treatment strategies, enhancing clinical decision-making and improving patient outcomes.The findings of this research contribute to the advancement of biomedical data classification by providing an effective, scalable, and computationally efficient methodology.

%============================= REFERENCE===============================
				

\bibliographystyle{unsrt}
\bibliography{ref}

\end{document}